\documentclass[lettersize,journal]{IEEEtran}
\usepackage{amsmath,amsfonts}
\usepackage{algorithmic}
\usepackage{algorithm}
\usepackage{array}
\usepackage[caption=false,font=normalsize,labelfont=sf,textfont=sf]{subfig}
\usepackage{textcomp}
\usepackage{stfloats}
\usepackage{xurl}
\usepackage{verbatim}
\usepackage{graphicx}
\usepackage{cite}
\usepackage{multirow}
\usepackage{booktabs}
\usepackage{color}
\usepackage[export]{adjustbox}

\usepackage{hyperref}

\begin{document}

\title{Balancing State Exploration and Skill Diversity in Unsupervised Skill Discovery}

\author{Xin Liu,~\IEEEmembership{Graduate Student Member,~IEEE,} Yaran Chen,~\IEEEmembership{Member,~IEEE,} Guixing Chen, 

Haoran Li,~\IEEEmembership{Member,~IEEE,} and Dongbin Zhao,~\IEEEmembership{Fellow,~IEEE}
        % <-this % stops a space

\thanks{This work is supported partly by the National Natural Science Foundation of China (NSFC) under Grants No. 62136008, and the International Partnership Program of the Chinese Academy of Sciences under Grant 104GJHZ2022013GC. \textit{(Corresponding author: Dongbin Zhao)}}
\thanks{X. Liu, Y. Chen, H. Li, and D. Zhao are with the State Key Laboratory
of Multimodal Artificial Intelligence Systems, Institute of Automation, Chinese Academy of Sciences, Beijing 100190, and also with the School of
Artificial Intelligence, University of Chinese Academy of Sciences, Beijing
100049, China. (email: liuxin2021@ia.ac.cn, chenyaran2013@ia.ac.cn, lihaoran2015@ia.ac.cn, dongbin.zhao@ia.ac.cn) Guixing Chen is with the Nanjing Institute of Software Technology. (email: guixing.chen@zwwlai.com)}
% <-this % stops a space
}

\markboth{Journal of \LaTeX\ Class Files, IEEE transactions on Cybernetics.}%
{Shell \MakeLowercase{\textit{et al.}}: A Sample Article Using IEEEtran.cls for IEEE Journals}


\maketitle

\begin{abstract}
    Unsupervised skill discovery seeks to acquire different useful skills without extrinsic reward via unsupervised Reinforcement Learning (RL), with the discovered skills efficiently adapting to multiple downstream tasks in various ways. However, recent advanced skill discovery methods struggle to well balance state exploration and skill diversity, particularly when the potential skills are rich and hard to discern. In this paper, we propose \textbf{Co}ntrastive dyna\textbf{m}ic \textbf{S}kill \textbf{D}iscovery \textbf{(ComSD)}\footnote{Code and videos: https://github.com/liuxin0824/ComSD} which generates diverse and exploratory unsupervised skills through a novel intrinsic incentive, named contrastive dynamic reward. It contains a particle-based exploration reward to make agents access far-reaching states for exploratory skill acquisition, and a novel contrastive diversity reward to promote the discriminability between different skills. Moreover, a novel dynamic weighting mechanism between the above two rewards is proposed to balance state exploration and skill diversity, which further enhances the quality of the discovered skills. Extensive experiments and analysis demonstrate that ComSD can generate diverse behaviors at different exploratory levels for multi-joint robots, enabling state-of-the-art adaptation performance on challenging downstream tasks. It can also discover distinguishable and far-reaching exploration skills in the challenging tree-like 2D maze. 
\end{abstract}


\section{Introduction \& Research Background}

\label{section-1}

\IEEEPARstart{U}{nder} the supervision of extrinsic rewards, Reinforcement Learning (RL) enables agents to acquire effective task-specific skills ~\cite{cyb-zhao,r14,rl-cyb2}. %The success of unsupervised learning in computer vision~\cite{simclr,swav} and natural language processing~\cite{nlp1,nlp2} further benefits task-specific RL with complex input~\cite{curl,drq,drq-v2,atc} by improving representation learning. 
%However, the agents trained with lots of efforts are always hard to generalize their knowledge to novel tasks due to the task-specific supervision of extrinsic reward~\cite{atc}. %Unsupervised RL is proposed to improve this generalization, where the task-agnostic reward is designed to conduct unsupervised pre-training. The pre-trained feature encoders and exploration policies then can be implemented for efficient RL on different downstream tasks. 
In contrast, however, real intelligent creatures, e.g., humans, are able to learn different useful task-agnostic skills without any supervision and utilize them to achieve multiple different tasks. To this end, unsupervised skill discovery is presented to simulate this process for machines~\cite{disdain,USD1-disk,usd2,lsd}. %As a branch of unsupervised RL ~\cite{exploration1,exploration2,exploration3}, skill discovery methods designs task-agnostic intrinsic rewards and achieves unsupervised pre-training with these rewards, which guarantees the multi-task downstream generalization. The main difference is that skill discovery requires extra input as conditions, named skill vectors. 
These approaches aim to generate various useful skills through training a skill-conditioned policy with a designed intrinsic reward, and the discovered skills can be utilized to efficiently achieve multiple downstream tasks in various ways, including skill finetuning, skill combination, and so on~\cite{diayn}. This necessitates (i) that the discovered skill set contains enough exploratory skills to ensure far-reaching exploration and state coverage (high state exploration) and (ii) that different skills corresponding to different skill condition vectors should be as different as possible (high skill diversity).

%~\citep{diayn}. %Designing proper unsupervised reward is the key to the success of skill discovery algorithms. 
%Currently, one of the most popular and effective classes of skill discovery is based on Mutual Information (MI) maximization~\cite{diayn,dads,aps,becl}. Most of them try to design intrinsic rewards to estimate and optimize the MI between skill vectors and visited states. With RL maximizing intrinsic reward expectation, correlations between skills and states are distilled, i.e., different useful skills are discovered. 

\begin{figure*}[t]
    \centering
    \includegraphics[width=0.93\textwidth]{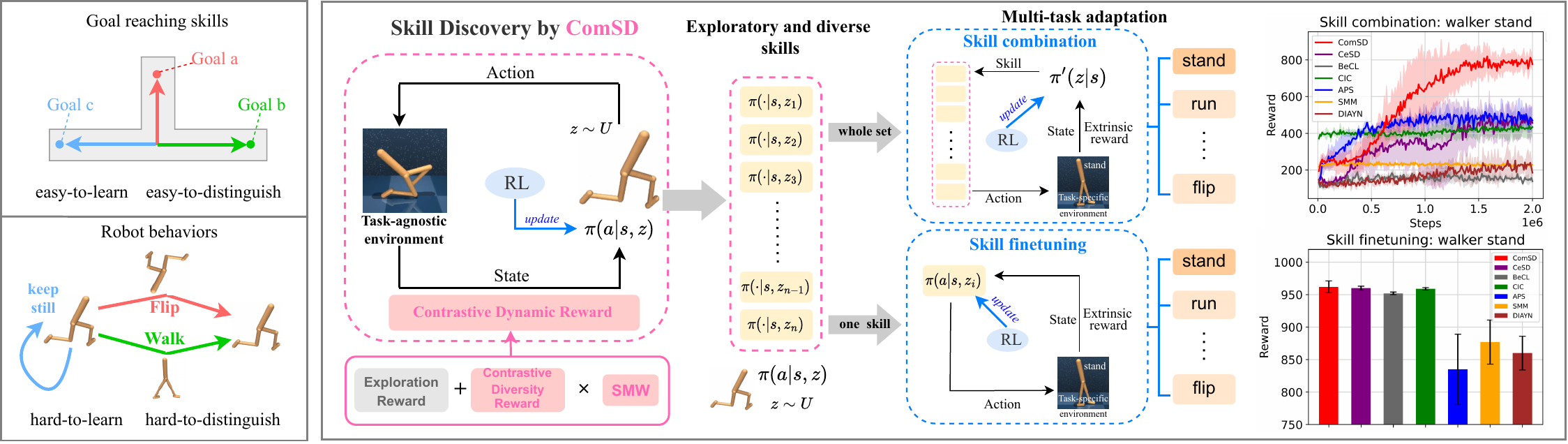}
    \vspace{-2mm}
    \caption{\textbf{Left}: The rich, hard-to-learn, and hard-to-distinguish potential skills significantly increase the difficulty of unsupervised skill discovery. First, the difficulty of learning exploratory skills, i.e., promoting state exploration, is increased. For example, learning to flip and walk for a robot is much harder than reaching two different goals due to the more complex robot kinetics. Second, the difficulty of discerning skills, i.e., promoting skill diversity, is increased. For example, reaching different goals can be easily represented by different ending states, while flipping and walking to the same location cannot be differentiated by only ending states. Third, behaviors at different exploratory levels are helpful for robots (e.g., static standing or dynamic running) but are not considered in goal reaching. \textbf{Right}: ComSD's pipeline. ComSD discovers exploratory and diverse robot behaviors through a novel contrastive dynamic reward, showing state-of-the-art performance on multiple downstream tasks in different kinds of evaluations.} %A detailed description of two adaptation tasks is shown in Section \ref{section-2-1}.} %Skill combination and skill finetuning together provide comprehensive adaptation evaluation, considering both behavioral exploration and diversity. A detailed description of two adaptation tasks is shown in Section \ref{section-2-1}. Our ComSD exhibits state-of-the-art adaptation performance on both kinds of downstream tasks, which recent advanced methods cannot. }
    \label{pipeline}
    \vspace{-3mm}
\end{figure*}

Currently, many unsupervised skill discovery methods have shown promising results in different environments, such as goal-reaching robots \cite{lsd,csd,metra} and maze exploration~\cite{becl,cesd}. The potential skill categories are relatively singular, and different skills are easily distinguishable in these domains. For more challenging benchmarks where the potential skills are rich and hard to distinguish by ending states, like URLB robot body movements~\cite{urlb}, the skill discovery difficulty is dramatically increased (see Fig. \ref{pipeline} left for a representative example). Recent advanced approaches cannot take both state exploration and skill diversity into account. %Learning to flip for a multi-joint robot is much harder than reaching a point in map exploration, and different robot behaviors are hard to distinguish by ending states while different reaching skills can be easily represented by ending states in map exploration. and  the recent advanced method cannot provide the exploration-diversity balance for the discovered behaviors. , .dy
For example, an ideal robot skill set should contain a variety of behaviors at different activity levels, while current advanced methods can either only learn different static postures (low state exploration)~\cite{diayn,aps} or only produce highly dynamic movements that are homogeneous (low skill diversity)~\cite{cic-nips,smm}. Some recent works have also noticed this issue~\cite{becl,cesd}, but they only alleviate it on maze exploration problems and can't well handle more challenging multi-joint robot behaviors. This is because their intrinsic rewards don't provide effective incentives to seek and discern rich body movements for complex robots, and they also ignore the conflict between state exploration and skill diversity (see Section V-I) when potential skills are hard to discern by ending states. %, enabling efficient downstream adaptation in various ways. 

In this paper, we propose \textbf{Co}ntrastive dyna\textbf{m}ic \textbf{S}kill \textbf{D}iscovery \textbf{(ComSD)} which seeks exploratory and diverse unsupervised skills through a novel contrastive dynamic reward. It contains strong incentives to separately encourage skill diversity and state exploration through contrastive learning, with a novel dynamic weighting proposed for their balance. Concretely, this reward aims to maximize state-skill mutual information (MI), which is decomposed into skill-conditioned entropy (related to skill diversity) and state entropy (related to state exploration). For skill-conditioned entropy optimization, we conduct state-skill contrastive learning, utilizing the contrastive results as a novel diversity reward. For state entropy maximization, we employ a particle-based estimation as the exploration reward. With contrastive learning, the above rewards achieve accurate entropy estimation, thus providing effective encouragement. In domains with hard-to-discern skills, we observe a serious mutually exclusive effect between the above two entropy optimization processes (see Section IV-C). Thus, a \textbf{S}kill-based dyna\textbf{M}ic \textbf{W}eighting (SMW) mechanism is proposed to weight the above rewards into our contrastive dynamic reward for balance between state exploration and skill diversity. We evaluate the adaptation ability of discovered robot behaviors on 16 skill combination downstream tasks and 16 skill finetuning downstream tasks (task details in Section III-A). ComSD enables state-of-the-art adaptation performance compared with five popular advanced methods, as shown in Fig. \ref{pipeline} right. A detailed behavioral quality analysis is also provided, demonstrating that ComSD indeed generates qualified robot behaviors of both high state exploration and high skill diversity, which recent advanced methods cannot. ComSD can also discover distinguishable and far-reaching exploration skills in the challenging tree-like 2D maze.

Our contributions can be summarized as follows: 
\begin{itemize}
    \item We propose \textbf{Co}ntrastive dyna\textbf{m}ic \textbf{S}kill \textbf{D}iscovery \textbf{(ComSD)}, a novel unsupervised skill discovery method, to generate exploratory and diverse unsupervised skills through a novel intrinsic incentive named contrastive dynamic reward. In this reward, we employ particle-based state entropy as an exploration reward for exploratory skill seeking, and propose to utilize the contrastive result between skill vectors and visited states as an explicit reward to promote diversity between different skills.
    \item To alleviate the exploration hurt caused by explicitly employing the contrastive diversity reward, we propose a novel weighting mechanism, named SMW, that dynamically scales the exploration reward and the diversity reward based on skill vectors. SMW is the key to well balancing state exploration and skill diversity, further improving the quality of discovered unsupervised skills.
    \item We conduct extensive numerical and analytical experiments. Results demonstrate that our proposed ComSD can discover diverse behaviors at different exploratory levels for challenging multi-joint robots, enabling state-of-the-art downstream adaptation ability on different kinds of downstream tasks. ComSD can also discover distinguishable and far-reaching exploration skills in the challenging tree-like 2D maze.
\end{itemize}

\section{Related Works}
Reinforcement Learning (RL) \cite{additional2,lihaoran,rl-cyb1} has proven its effectiveness in learning useful task-specific skills in the presence of extrinsic rewards~\cite{r44,rl-cyb3}. The success of unsupervised learning in computer vision~\cite{simclr} and natural language processing~\cite{nlp1} further benefits task-specific RL with visual input~\cite{curl,drq-v2,atc} by improving representation learning. However, the agents trained with lots of efforts are always hard to generalize their knowledge to novel tasks due to the task-specific supervision of extrinsic reward~\cite{atc}. Inspired by the reward shaping approaches employed in inverse RL~\cite{additional1,IRL,irl-cybsurvey}, unsupervised RL~\cite{urlb} is proposed to improve the multi-task generalization by designing task-agnostic rewards for unsupervised pre-training. The pre-trained feature encoders~\cite{apt,proto-rl,crptpro} and exploration policies~\cite{exploration2,exploration3,exploration20024-1} then can be implemented for efficient RL on different downstream tasks. 
In addition to the generalization, intelligent agents should also be able to explore environments and learn different useful skills without any extrinsic supervision, like human beings. For the reasons above, unsupervised skill discovery~\cite{disdain,url-theory} is proposed and becomes a novel research hotspot. As a branch of unsupervised RL, unsupervised skill discovery also designs task-agnostic rewards and achieves unsupervised pre-training with these rewards, which guarantees task-agnostic downstream generalization\cite{usd2024-1,usd2024-2}. The main difference is that skill discovery requires extra input as conditions, named skill vectors. They aim to discover useful task-agnostic policies that are distinguishable by skill vectors. These approaches exhibit promising results across different fields~\cite{gym-env,atari-env,map-exploration-env}. However, in the URLB benchmark \cite{urlb} where the potential skills are rich and hard to distinguish by ending states (discussion in Fig. \ref{pipeline} left), it's hard for current advanced MI-based methods \cite{cic-nips,becl,cesd} to take both state exploration and skill diversity into account. Recently, the distance-maximizing skill discovery methods \cite{lsd,csd,metra} are proposed to discover diverse and far-reaching exploration skills for challenging goal-reaching tasks. However, these advanced goal-reaching skills are not very suitable for solving different types of the URLB downstream tasks that require different kinds of body movements \cite{cesd}. In this paper, ComSD designs a dynamically weighted intrinsic incentive based on contrastive learning, discovering diverse robot behaviors at different exploratory levels in the URLB benchmark, which recent advanced methods cannot achieve.

%In addition to unsupervised skill discovery, Quality-Diversity (QD) optimization, a novel type of evolution algorithm, is also proposed to generate diverse behaviors \cite{qd-first}. With two functions to respectively judge the novelty and quality of behaviors, QD methods \cite{qd1,qd2,qd3} can produce large collections of diverse solutions that are high-performing. However, it's hard to define proper quality functions or novelty functions for robot behavior discovery. For example, we can't use speed as the 'quality' because static postures are also useful (e.g., walker stand) for robots. We also can't judge 'novelty' by the ending location because walking and flipping to one point are not the same. These also further demonstrate how challenging robot behavior discovery is. In addition, as evolution algorithms, QD approaches can only discover limited solutions. By contrast, our ComSD benefits from generalized deep learning, being able to discover infinite skills through a continuous skill space.

\section{Preliminaries}
\subsection{Problem Definition}
\label{section-2-1}

Unsupervised skill discovery %algorithms aim at discovering a useful and diverse set of agent behaviors in the absence of extrinsic reward, i.e., 
trains a skill-conditioned policy in a task-agnostic environment. %The learned skills are then utilized for downstream task-specific policy learning with extrinsic reward. %Therefore, there are usually two phases (task-agnostic pre-training stage and task-specific skill utilization stage) related to different kinds of Markov Decision Process (MDP).
Concretely, a reward-free Markov Decision Process (MDP)~\cite{mdp} is considered and defined as $\mathcal{M}^{free}=(\mathcal{S},\mathcal{A},\mathcal{P},\gamma,d_0)$, where $\mathcal{S}$ is the state space, $\mathcal{A}$ is the action space, $\mathcal{P}$ is the distribution of the next state given the current state and action, $\gamma$ is the discount factor, and $d_0$ is the distribution of the initial state. What skill discovery algorithms do is define an intrinsic reward $r^{intr}$ and augment $\mathcal{M}^{free}$ to an intrinsic-reward MDP $\mathcal{M}^{intr}=(\mathcal{S},\mathcal{A},\mathcal{P},r^{intr},\gamma,d_0)$. Then we denote the skill space by $\mathcal{Z}$. With the skill vector $z$ ($z\in \mathcal{Z}$) sampled from the skill distribution $p(z)$ defined by skill discovery algorithms, a skill-conditioned policy $\pi(a|s,z)$ can be obtained by RL over the $\mathcal{M}^{intr}$. We provide the pseudo-code of our ComSD in Algorithm \ref{algorithm-comsd} as a representative example to show the general pipeline of unsupervised skill discovery.

To evaluate the quality of discovered skills (i.e., the skill-conditioned agent $\pi(a|s,z)$), two adaptation evaluations: skill combination and skill finetuning are employed on each task-specific downstream task. A downstream task can be described as an extrinsic-reward MDP $\mathcal{M}^{extr}=(\mathcal{S},\mathcal{A},\mathcal{P},r^{extr},\gamma,d_0)$, where $r^{extr}$ denotes the task-specific extrinsic reward. In skill combination, the learned $\pi(a|s,z)$ is frozen, and a meta-controller $\pi'(z|s)$ is required to choose and combine skill vectors for $\pi(a|s,z)$ automatically on downstream tasks. Concretely, over the original downstream task $\mathcal{M}^{extr}$, skill combination can be defined as $\mathcal{M}^{extr'}=(\mathcal{S},\mathcal{Z},\mathcal{P'},r^{extr'},\gamma,d_0)$, which regards different skill vectors $z \sim \mathcal{Z}$ as its actions and correspondingly changes its transition model and reward function. The meta-controller $\pi'(z|s)$ is trained over $\mathcal{M}^{extr'}$ by RL and then used for evaluation of the learned $\pi(a|s,z)$. In skill finetuning, a target skill vector $z_i$ is chosen. The corresponding policy $\pi(a|s,z_i)$ is further finetuned on the $\mathcal{M}^{extr}$ for a few steps and serves as the evaluation of the discovered $\pi(a|s,z)$. The pseudo-codes of two adaptation evaluations are provided in Algorithm \ref{algorithm-combination} and Algorithm \ref{algorithm-finetuning} respectively (Section V).

\subsection{Mutual Information Objective of Unsupervised RL}
\label{section-2-2}

As current state-of-the-art approaches do~\cite{cic-nips,aps,diayn}, our ComSD tries to maximize the Mutual Information (MI) objective between state function $\tau$ and skill vector $z$ for behavior discovery. %In general, $\tau(s)$ can be (i) maintaining the original state $\tau(s) = s$, (ii) concatenating the neighboring state transitions $\tau(s) = {\rm concat}(s^{t-1},s^{t} )$, or (iii) using the whole trajectory $\tau(s) = {\rm concat}(s^1,...,s^{t})$. %Following the recent advanced method~\cite{cic-nips}, 
$\tau$ is defined as state transitions ($\tau = {\rm concat}(s^{t-1},s^{t})$) 
throughout this paper. The MI objective $I(\tau;z)$ can be decomposed into the form of Shannon entropy in two ways:

\begin{equation}
\label{eq1}
    I(\tau;z) = -H(z|\tau) + H(z),
\end{equation}
\begin{equation}
\label{eq2}
    I(\tau;z) = -H(\tau|z) + H(\tau),
\end{equation}

\noindent where the $I(\cdot;\cdot)$ denotes the MI function and $H(\cdot)$ denotes the Shannon entropy throughout the paper. Most classical MI-based algorithms~\cite{vic,diayn,smm} are based on the first decomposition Eq. (\ref{eq1}). A uniform random policy can guarantee the maximum of the skill entropy $H(z)$ while the trainable discriminators estimating the state-conditioned entropy $H(z|\tau)$ are employed as the intrinsic rewards for unsupervised RL. However, ~\cite{urlb,cic-nips} have found it hard for these methods to guarantee robot state exploration without any external balancing signals. To this end, some recent works~\cite{dads,aps,cic-nips} focus on the second decomposition Eq. (\ref{eq2}) to explicitly optimize the state entropy $H(\tau)$ by designing exploration rewards, achieving considerable performance in different domains, including robotic manipulation, video games, and challenging robot locomotion. Following these, we choose Eq. (\ref{eq2}) as our optimization target.

\begin{figure}[t]
    \centering
    \includegraphics[width=0.4\textwidth]{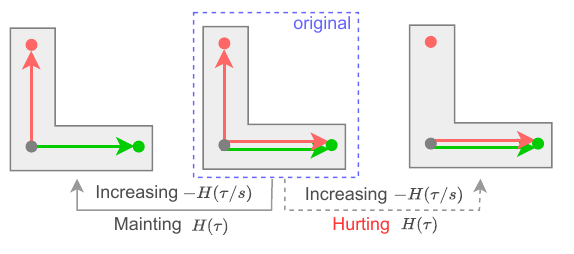}
    \vspace{-2mm}
    \caption{ An example to illustrate why reducing exploration within each skill (increasing $-H(\tau|z)$) with the overall exploration of all skills ($H(\tau)$) guaranteed can increase the skill diversity. The red skill reaches both the red goal and the green goal originally, while the green skill reaches only the green goal. Here we ignore trajectories, treating different goals as different states. Increasing $-H(\tau|z)$ forces the red skill to give up one of the two goals for a lower own state exploration. If the red skill reaches the green goal like the green skill (right in the figure), no skill can reach the red goal, and the overall state coverage $H(\tau)$ will be reduced. To this end, the only way to increase $-H(\tau|z)$ without hurting $H(\tau)$ is that the red skill reaches the red goal (left in the figure), which means a larger difference between two skills.}
    \label{entropy}
\end{figure}

\section{Unsupervised Skill Discovery by ComSD}

\begin{figure*}[t]
    \centering
    \includegraphics[width=0.8\textwidth]{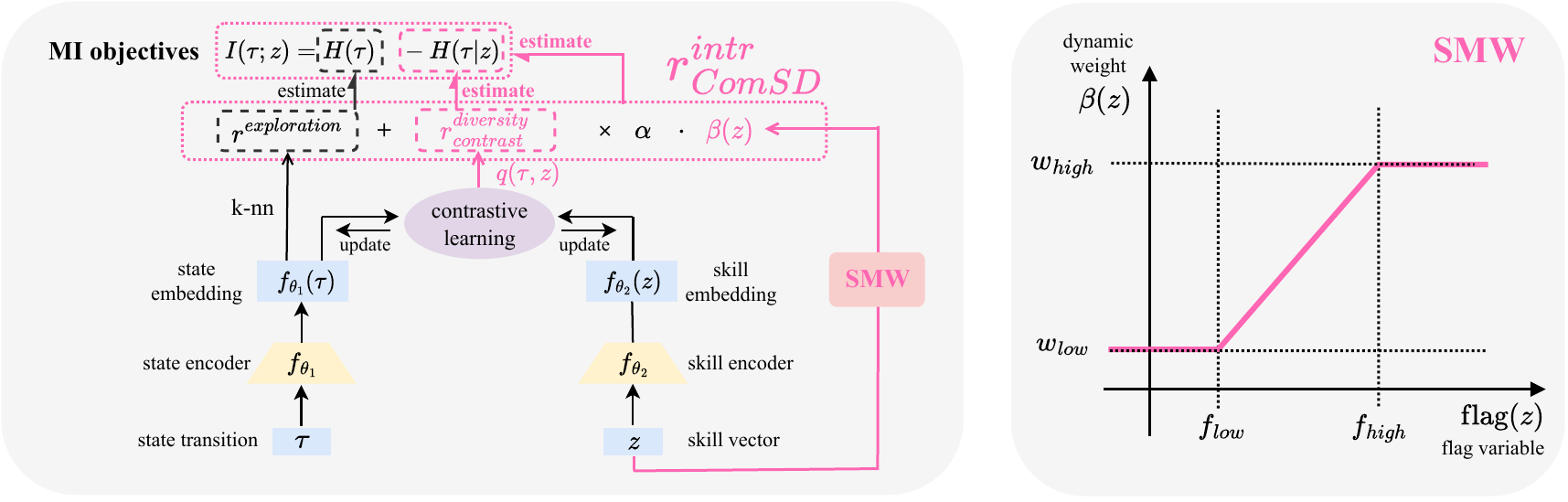}
    \caption{\textbf{Left}: The contrastive dynamic reward $r^{intr}_{ComSD}$ design. \textbf{Right}: In SMW, $\beta$ is linear related to $\rm flag \it(z)$ with different slopes in different region. The skill space is divided into different regions with different learning objectives.}
    \label{intrinsicreward}
\end{figure*}

In this section, we show how ComSD seeks skills of both skill diversity and state exploration via a novel contrastive dynamic reward $r^{intr}_{ComSD}$ based on Eq. (\ref{eq2}). In Section IV-A, we propose a novel diversity reward $r^{diversity}_{contrast}$ through contrastive learning to estimate the skill-conditioned entropy. In Section IV-B, we illustrate the way to estimate the state entropy by a particle-based exploration reward $r^{exploration}$. In Section IV-C, we detail why and how to balance the above two rewards into our contrastive dynamic reward with the proposed SMW. The overall design of our proposed contrastive dynamic reward $r^{intr}_{ComSD}$ is shown in Fig. \ref{intrinsicreward}.

\subsection{Skill-conditioned Entropy Estimation via Contrastive Learning}
\label{section-3-1}

In decomposition Eq. (\ref{eq2}), higher negative conditioned entropy $-H(\tau|z)$ means lower state coverage within each skill. When the overall exploration of all skills, i.e., state entropy $H(\tau)$, is guaranteed (discussed in Section IV-B), increasing $-H(\tau|z)$ reduces exploration within each skill, forcing the agent to exploit skill-specific information and maintain unique information for different skills, i.e., increasing the diversity of learned skills. We provide an example to illustrate this process, as shown in Fig. \ref{entropy}.

Due to the unavailable conditioned distribution, we can't directly maximize the $-H(\tau|z)$ but try to maximize its lower bound. First, $-H(\tau|z)$ can be decomposed as:

\begin{equation}
\begin{aligned}
-H(\tau|z) &= \sum_{\tau,z}p(\tau,z)\log p(\tau|z) \\
&= \sum_{\tau,z}p(\tau,z)\log \frac{p(\tau,z)}{p(z)}\\
&= \sum_{\tau,z}p(\tau,z)(\log  p(\tau,z)-\log p(z)),
\end{aligned}
\end{equation}

\noindent where $p(\cdot)$ denotes probability throughout the paper. Following previous works \cite{becl,cic-nips}, we use a uniform distribution to generate the skill vectors. %, which follows recent works ~\citep{diayn,smm,cic-nips,becl}. 
Therefore, the term $\rm log\it p(z)$ becomes a constant, and we denote it as $c$:

\begin{equation}
\begin{aligned}
-H(\tau|z)&= \sum_{\tau,z}p(\tau,z)(\rm log\it p(\tau,z)-c) \\
&= \sum_{\tau,z}p(\tau,z)\rm log\it p(\tau,z)- \sum_{\tau,z}p(\tau,z) \cdot c\\
&= \sum_{\tau,z}p(\tau,z)logp(\tau,z) - c. %\propto \sum_{\tau,z}p(\tau,z)\log p(\tau,z). %\\
%&\geq \sum_{\tau,z}p(\tau,z)logp(\tau,z).
\end{aligned}
\end{equation}

We neglect $c$ and then obtain the variational lower bound:

\begin{equation}
\begin{aligned}
-H'(\tau|z) & = \sum_{\tau,z}p(\tau,z)\log p(\tau,z)\\
&=\sum_{\tau,z}p(\tau,z)\log p(\tau,z)  \\
&+ \sum_{\tau,z}p(\tau,z)\log q(\tau,z) -\sum_{\tau,z}p(\tau,z)\log q(\tau,z)\\
&=\sum_{\tau,z}p(\tau,z)\log q(\tau,z) +D_{\rm KL}(p(\tau,z)||q(\tau,z)) \\
&=\mathbb{E}_{\tau,z}[\log q(\tau,z)] +D_{\rm KL}(p(\tau,z)||q(\tau,z)) \\
&\ge \mathbb{E}_{\tau,z}[\log q(\tau,z)],
\end{aligned}
\end{equation}

\noindent where $q(\tau,z)$ is employed to estimate the unavailable $p(\tau,z)$. To optimize $-H(\tau|z)$, we need to minimize the KL divergence $D_{\rm KL}(p(\tau,z)||q(\tau,z))$ and maximize the expectation $\mathbb{E}_{\tau,z}[\log q(\tau,z)]$. 

For KL divergence minimization, the distribution $q$ should be as positively correlated with the probability function $p$ as possible. We define $q$ as the exponential inner product between the state embeddings and skill embeddings:

\begin{equation}
    q(\tau,z) = \exp\frac{f_{\theta_1}(\tau)\cdot f_{\theta_2}(z)}{||f_{\theta_1}(\tau)||\cdot||f_{\theta_2}(z)||T},
\end{equation}

\noindent where $f_{\theta_1}(\cdot)$ is the state encoder, $f_{\theta_2}(\cdot)$ is the skill encoder, $T$ is the temperature parameter, and $\exp(\cdot)$ makes $q$ non-negative like $p$. The above two encoders are updated by gradient descent after computing the NCE loss~\cite{nce} in contrastive learning~\cite{simclr}:
\begin{equation}
\label{eq7}
    \mathcal{L}_{NCE} = \log q(\tau_i,z_i)-\log \frac{1}{N} \sum_{j=1}^N q(\tau_j,z_i),
\end{equation}

\noindent where $\tau_i$ is sampled by skill $z_i$ and $\tau_j$ is sampled by other skills. In most circumstances, the probability $p(\tau_i,z_i)$ is much larger than $p(\tau_j,z_i)$. Maximizing $\mathcal{L}_{NCE}$ increases the value of $q(\tau_i,z_i)$ and decreases the value of $q(\tau_j,z_i)$, which actually shrinks the difference between $p$ and $q$.

For expectation maximization, we regard the content (the contrastive result) as a part of the intrinsic reward, employing RL to optimize it:

\begin{equation}
\label{eq8}
    r^{diversity}_{contrast} \propto q(\tau_i,z_i),
\end{equation}

\noindent where $r^{diversity}_{contrast}$ is our contrastive diversity reward.

\subsection{Particle-based State Entropy Estimation}
\label{section-3-2}

In decomposition Eq. (\ref{eq2}), increasing the state entropy $H(\tau)$ encourages the agent to explore more widely and visit more state transitions. Following ~\cite{aps,proto-rl,cic-nips}, we employ a particle-based entropy estimation proposed by ~\cite{apt}. Concretely, $H(\tau)$ can be estimated by the Euclidean distance between each particle ($\tau$ in our paper) and its all $k$-nearest neighbor~\cite{nn-estimate} in the latent space:

\begin{equation}
    H(\tau) \approx H_{particle}(\tau) \propto \frac{1}{k}\sum_{h_i^{nn}\in N_{buffer}^{k-nn}(h_i)} \log||h_i-h_i^{nn}||,
\end{equation}

\noindent where $h_i$ can be any forms of encoded $\tau_i$ and $N_{buffer}^{k-nn}(h_i)$ denotes neighbor set. Following ~\cite{cic-nips}, we choose the state encoder in contrastive representation learning and calculate this entropy over sampled RL training batch. The exploration reward is defined as the following:

\begin{equation}
\label{eq10}
\begin{aligned}
    r^{exploration} \propto \frac{1}{k}\sum_{h_i^{nn}\in N_{batch}^{k-nn}(h_i)}  \log||h_i&-h_i^{nn}||, \\
   where \  \ h_i &= f_{\theta_1}(\tau_i).
\end{aligned}
\end{equation}

\subsection{\textbf{S}kill-based Dyna\textbf{M}ic \textbf{W}eighting (SMW)}
\label{section-3-3}

With the negative conditioned entropy $-H(\tau|z)$ estimated by $r^{diversity}_{contrast}$ and the state entropy estimated by $r^{exploration}$, a naive way for intrinsic reward design is to employ a fixed coefficient $\alpha$ to scale the two terms ($r^{intr} = r^{exploration} +\alpha \cdot r^{diversity}_{contrast}$). However, in practice, we find it impossible to use a fixed coefficient to balance them well when potential skills are rich, hard to explore, and hard to distinguish by only ending states. Concretely, when employing $r^{exploration}$ alone, the robots can only learn dynamic movements of high activity but not other exploratory levels, and the learned behaviors are highly homogeneous (high exploration but low exploitation). Increasing the fixed weight $\alpha$ of $r^{diversity}_{contrast}$ can surely improve the skill diversity, but it will simultaneously reduce the state exploration (see experiments in Section V-I), making robots prefer static postures (easily distinguished by simple ending states) but ignore active movements (distinguished by trajectories but not only ending states). The large fixed weight leads to high exploitation but low exploration. This is because distilling ending states into skill vectors is an easy-to-reach local optimal solution for state-skill MI maximization. In addition, this conflict between exploration and diversity also exists in other advanced methods (Section V-I).

We propose \textbf{S}kill-based dyna\textbf{M}ic \textbf{W}eighting (SMW), an easy-but-effective weighting mechanism to mitigate the issue above. The main idea of SMW is to set different learning objectives for different skills instead of using fixed learning objectives across all skills. Concretely, it designs another changeable coefficient $\beta(z)$ which is dynamically adjusted according to skill vectors. $\beta(z)$ is defined as:

\begin{equation}
\begin{aligned}
& \beta(z) = \rm clamp \it ( \beta', (w_{high},w_{low})), \\
where \  \ \beta' & = (\frac{w_{high}-w_{low}}{f_{high}-f_{low}})(\rm flag \it (z)-f_{high})+w_{high}.
\end{aligned}
\end{equation}

The $\rm flag \it(\cdot)$ is a function to map the high-dimensional skill vectors into 1-dimensional flag variables. Here, we find that naively employing the first dimension of skill vectors as the flag variables (i.e., $\rm flag \it(z) = I_1^T\cdot z,\ {\rm where} \ I_1 = (1,0,...,0)$) enables considerable results in practice. The fixed hyper-parameters $w_{high},w_{low},f_{high},f_{low}$ and the clamp function $\rm clamp \it(\cdot)$ make the dynamic coefficient $\beta$ linearly related to flag variables $\rm flag \it (z)$ with different slopes in different regions. We clearly visualize the relation between $\beta$ and $\rm flag \it(z)$ in Fig. \ref{intrinsicreward} right. With SMW, the skill space is actually divided into different regions with different learning objectives. In the high-exploration region ($\rm flag \it(z) <f_{low} $, corresponding to low diversity reward weight), the agents try their best to explore, while in the high-exploitation region ($\rm flag \it(z) >f_{high} $, corresponding to high diversity reward weight), the agents pay more attention to discerning learned skills. In the middle ($\rm flag \it(z)$ between $f_{low}$ and $f_{high}$), the agents can acquire behaviors at different activity levels (corresponding to different $\beta$ values). The contrastive dynamic reward of ComSD is defined as:

\begin{equation}
\label{eq12}
    r^{intr}_{ComSD} = r^{exploration} + \alpha \cdot \beta(z) \cdot r^{diversity}_{contrast}.
\end{equation}

Over the intrinsic-reward MDP $\mathcal{M}^{intr}$ augmented by $r^{intr}_{ComSD}$, ComSD trains the skill-conditioned policy $\pi(a|s,z)$ via RL with $z$ sampled by uniform distribution. Algorithm \ref{algorithm-comsd} provides the full pseudo-code of ComSD.

\begin{algorithm}[t]
\caption{Pseudo-code of ComSD.}
%\textsc{\#\#\#Unsupervised skill discovery by ComSD}

\textbf{Require:} Reward-free environment $E_f$, the uniform skill distribution $p(z)$, unsupervised pre-training environment steps $I_p$, and the RL batch size $I_b$.

\textbf{Initialize:} The state encoder $f_{\theta_1}(\cdot)$, the skill encoder $f_{\theta_2}(\cdot)$, the skill-conditioned policy (actor) $\pi(a|s,z)$, the critic $Q(a,concat(s,z))$, and the replay buffer $D$.

\begin{algorithmic}[1]

\STATE \hspace{0cm}\textbf{for} $t = 1,...,I_p $ \textbf{do}
\STATE \hspace{0.5cm}Sample a skill vector $z_t$ from uniform distribution.
\STATE \hspace{0.5cm}Obtain current action $a_{t} \sim \pi(a|s_t,z_t)$ based on current state $s_t$ and skill vector $z_t$. 
\STATE \hspace{0.5cm}Interact with $E_f$ with $a_t$ to get next state $s_{t+1}$.

\STATE \hspace{0.5cm}Add the transition $(s_t,z_t,a_t,s_{t+1})$ into $D$.
\STATE \hspace{0.5cm}Sample $I_b$ transitions from $D$ after enough collections.
%\STATE \hspace{0.5cm}Encode state transitions $\tau(s) = concat(s^{t-1},s^{t})$ by state encoder $f_{\theta_1}(\cdot)$.
%\STATE \hspace{0.5cm}Encode the skill vector $z_t$ by skill encoder $f_{\theta_2}(\cdot)$.

\#\#\#\# \textit{ComSD contrastive learning (line 7-8)} \#\#\#\#
\STATE \hspace{0.5cm}Compute $\mathcal{L}_{NCE}$ in Eq. (\ref{eq7}) with $f_{\theta_1}(\cdot)$ and $f_{\theta_2}(\cdot)$.
\STATE \hspace{0.5cm}Use backpropogation to update $f_{\theta_1}(\cdot)$ and $f_{\theta_2}(\cdot)$.

\#\#\#\# \textit{ComSD reward computation (line 9-11)} \#\#\#\#
\STATE \hspace{0.5cm}Compute diversity reward $r^{diversity}_{contrast}$ in Eq. (\ref{eq8}).
\STATE \hspace{0.5cm}Compute exploration reward $r^{exploration}$ in Eq. (\ref{eq10}).
\STATE \hspace{0.5cm}Compute intrinsic reward $r^{intr}_{ComSD}$ in Eq. (\ref{eq12}).

\STATE \hspace{0.5cm}Augment the sampled $I_b$ transitions by $r^{intr}_{ComSD}$.
\STATE \hspace{0.5cm}Use DDPG to update $\pi(a|s,z)$ and $Q(a,concat(s,z))$ over $I_b$ intrinsic-reward transitions.

\STATE \hspace{0cm}\textbf{end for}
\end{algorithmic}
\textbf{Output:} the discovered skills (trained skill-conditioned policy) $\pi(a|s,z)$.

\label{algorithm-comsd}
\end{algorithm}

\section{Experiments \& Analysis}
\label{section-4-1}

\subsection{Environments} 

\textit{On the importance of environments:} OpenAI Gym~\cite{gym-env} and DMC~\cite{dmc} are two of the most popular continuous robot locomotion benchmarks. In Gym, the episode ends when agents lose their balance, while the episode length in DMC is fixed. \cite{cic-nips} found this small difference makes DMC much harder for reward-free exploration, and most classical methods succeeding in OpenAI Gym suffer from lazy exploration in DMC. This is mainly because the agents have to learn balance by themselves without any external signals in DMC. To this end, we employ the challenging DMC~\cite{dmc} and URLB \cite{urlb} (the most popular skill discovery benchmark based on DMC) as our environments.  

Following recent advanced methods~\cite{cic-nips,proto-rl,crptpro,curl,atc,drq-v2}, we employ 16 downstream tasks of 4 domains from URLB~\cite{urlb} and DMC~\cite{dmc} for skill adaptation evaluation. The domains Walker, Quadruped, Cheetah, and Hopper are recognized as the most representative and challenging multi-joint robot environments, each of which contains 4 totally diverse and challenging downstream tasks.

In each domain, all the methods pre-train their agents for 2M environment steps with their respective intrinsic rewards, which follows previous works \cite{cic-nips,becl}. The hyper-parameters of all baselines follow the official implementations \cite{urlb} and we provide the detailed settings of ComSD in Table \ref{hyper}. After unsupervised pre-training, the discovered skills will be fully evaluated on two adaptation evaluations: skill combination~\cite{diayn} and skill combination in URLB~\cite{urlb} respectively, across all 16 downstream tasks with extrinsic reward. It means a total of 32 numerical results are employed for one method's evaluation. DDPG~\cite{ddpg} is chosen as the backbone RL algorithm for all methods throughout the paper. The detailed settings of the skill combination and skill finetuning are provided along with the experimental results. 

Following recent advanced works \cite{becl,cesd}, we also conduct experiments on 2D maze exploration \cite{map-exploration-env}, where an agent is required to explore the different parts under different skills and maximize its whole exploration in the maze. Different skills can be well represented by the sampled trajectories that can be easily visualized by different colors. We choose the challenging tree-like maze, which consists of two branches and four leaves. See Section V-K for concrete maze structure. The environment settings all follow previous works \cite{becl,cesd}.

\subsection{Baselines}

We compare our ComSD with six recent advanced algorithms: CeSD \cite{cesd}, BeCL~\cite{becl}, CIC~\cite{cic-nips}, APS~\cite{aps}, SMM~\cite{smm}, and DIAYN~\cite{diayn}. These advanced methods are currently state-of-the-art for challenging robot behavior discovery \cite{urlb} and maze exploration \cite{map-exploration-env}.

DIAYN~\cite{diayn} is one of the most classical and original unsupervised skill discovery algorithms, proposing to maximize the MI between skills and states. It employs the first MI decomposition, Eq. (\ref{eq1}). It uses a discrete uniform prior distribution to guarantee the maximization of skill entropy $H(z)$. The negative state-conditioned entropy $-H(z|s)$ is estimated by a trainable discriminator ${\rm log}p(z|s)$ which computes the intrinsic reward. As a foundational work, it provides several reasonable evaluations of skill adaptation, of which skill finetuning and skill combination are employed in our experiments. 

SMM~\cite{smm} aims to learn a policy for which the state marginal distribution matches a given target state distribution. It optimizes the objective by reducing it to a two-player, zero-sum game between a state density model and a parametric policy. Like DIAYN, it is also based on the first decomposition Eq. (\ref{eq1}) of the MI objective.

\begin{algorithm}[b]
\caption{Pseudo-code of skill combination evaluation.}

%\textsc{\#\#\#Adaption evaluation of discovered skills by skill combination}

\textbf{Require:} Reward-specific environment $E_s$, the pre-trained skill-conditioned policy $\pi(a|s,z)$, environment adaption steps $I_a$, and the RL batch size $I_b$.

\textbf{Initialize:} The meta-controller (actor) $\pi'(z|s)$, the critic $Q(z,s)$, and the replay buffer $D$.

\begin{algorithmic}[1]
\STATE \hspace{0cm}Freeze the learned skills $\pi(a|s,z)$.
\STATE \hspace{0cm}\textbf{for} $t = 1,...,I_a $ \textbf{do}
%\STATE \hspace{0.5cm}Sample a skill vector from uniform distribution $z_t \sim p(z)$.
\STATE \hspace{0.5cm}Obtain the current skill vector $z_{t} \sim \pi'(z|s_t)$. $s_t$ is the current state. 
\STATE \hspace{0.5cm}Obtain action $a_{t} \sim \pi(a|s_t,z_t)$. 
\STATE \hspace{0.5cm}Interact with reward-specific environment $E_s$ with $a_t$ to get next observation $s_{t+1}$ and the extrinsic reward $r^{extr}$.
\STATE \hspace{0.5cm}Add the transition $(s_t,z_t,r^{extr},s_{t+1})$ into $D$.
\STATE \hspace{0.5cm}Sample $I_b$ transition batch from $D$.

\STATE \hspace{0.5cm}Use DDPG to update $\pi'(z|s)$ and $Q(z,s)$ over $I_b$ transitions.

\STATE \hspace{0cm}\textbf{end for}
\end{algorithmic}

\textbf{Output:} The performance of $\pi'(z|s)$ serves as the skill combination evaluation result.

\label{algorithm-combination}
\end{algorithm}

APS~\cite{aps} employs the second decomposition Eq. (\ref{eq2}) for its MI estimation. For state entropy estimation, it employs a popular particle-based entropy estimation proposed by APT~\cite{apt} for exploration. For skill-conditioned entropy, it chooses the successor feature~\cite{visr} as a diversity reward with a fixed weight. APS achieves considerable results on video games \cite{atari-env} but can't provide enough incentives for robots, suffering from lazy exploration \cite{cic-nips}. 

CIC~\cite{cic-nips} is a state-of-the-art robot behavior discovery method on URLB. It first introduces contrastive learning~\cite{simclr} into unsupervised skill discovery. It chooses the second MI decomposition, Eq. (\ref{eq2}) with APT particle-based estimation for state entropy estimation, like APS. The contrastive learning between state transitions and skill vectors is conducted for implicit skill-conditioned entropy maximization. We find that the robot behaviors produced by CIC are of high activity but low skill diversity. 

BeCL~\cite{becl} is another state-of-the-art method on URLB~\cite{urlb} and maze exploration~\cite{map-exploration-env}. It tries to mitigate the diversity problem in CIC by designing a novel MI objective, $I(s^1,s^2)$, where $s^1$ and $s^2$ denote different states generated by the same skill. BeCL can't generate enough dynamic robot behaviors, and their intrinsic reward computational consumption is also much larger than other approaches.

CeSD \cite{cesd} uses an ensemble of skills for partitioned exploration based on state prototypes, enabling local exploration within clusters while maximizing overall state coverage. It also employs the second decomposition Eq. (\ref{eq2}) for MI estimation. It achieves advanced results across both robot locomotion \cite{urlb} and maze explorations~\cite{map-exploration-env}.

The close baselines to our ComSD are CIC and APS, where they all design intrinsic rewards based on Eq. (\ref{eq2}). Apart from the most notable difference, a novel weighting mechanism (SMW), ComSD still significantly differs from these methods. CIC doesn't design a diversity incentive, while ComSD proposes a novel contrastive diversity reward to explicitly encourage diversity. APS also employs both diversity reward and exploration reward, but it doesn't conduct contrastive learning that is the key to providing effective and strong incentives for challenging robots. In addition, BeCL also employs contrastive learning, but their contrastive components are different states generated by the same skill, which differs from ours.  

%\section{Experiments: Adaptation Results}
%\label{section-4-2}

\begin{figure*}[t]
    \centering
    \includegraphics[width=0.9\textwidth]{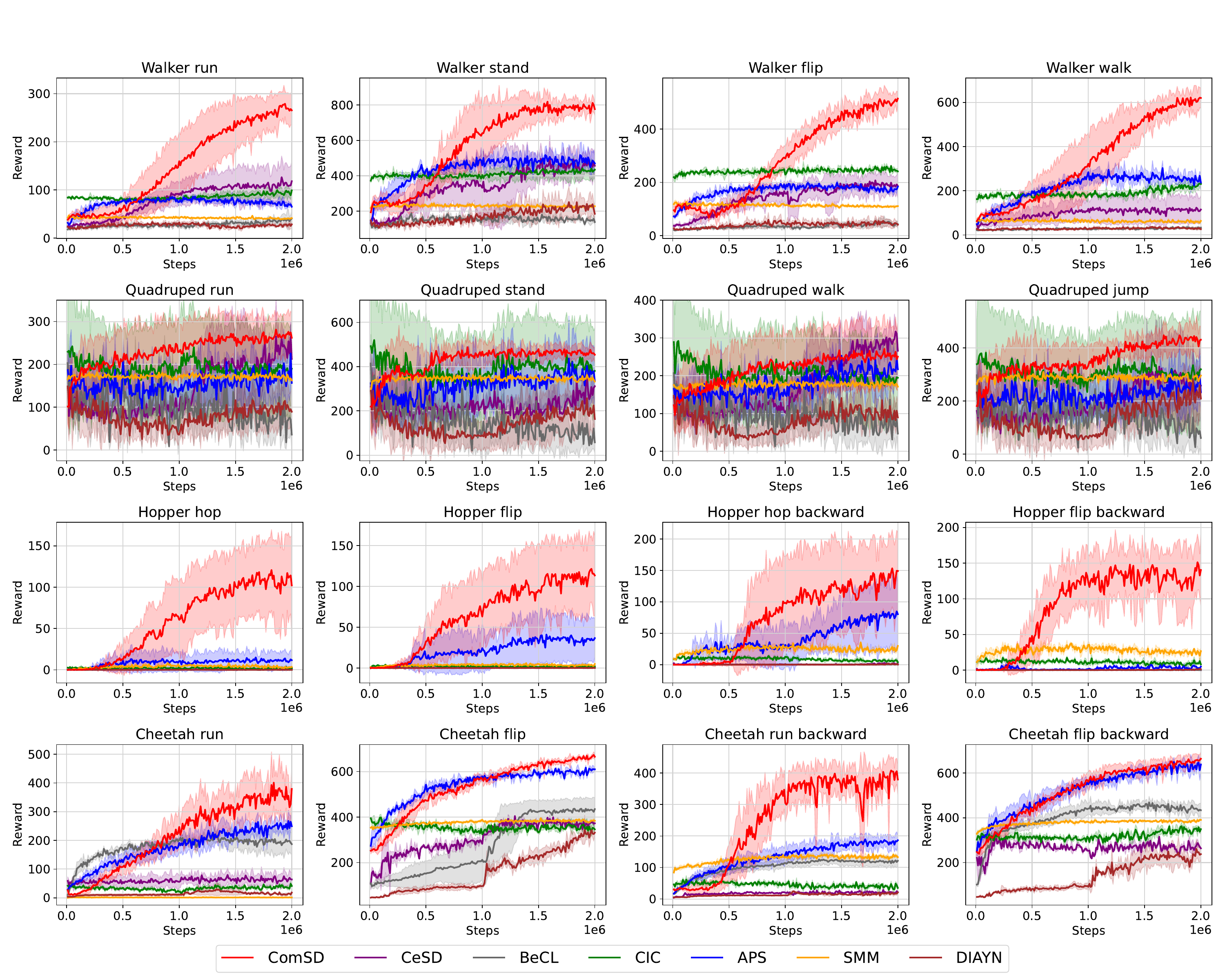}
    \caption{The training curve of all 7 methods on 16 skill combination downstream tasks. ComSD outperforms all 6 baselines significantly across 15/16 downstream tasks, demonstrating that ComSD discovers much more exploratory and diverse behaviors than other methods for challenging multi-joint robots.}
    \label{skillcombination}
\end{figure*}

\begin{table*}[t]

\centering
\renewcommand\arraystretch{1.1}
\caption{The numerical results of all 7 methods on 16 skill finetuning downstream tasks. Considering the RL high variance, we bold the highest result and the results with a small interval (2\% of the full score 1000, i.e., 20) throughout this paper. %Despite the adverse circumstances (much higher skill diversity leads to much harder skill choice), 
ComSD exhibits competitive adaptation performance compared with state-of-the-art methods (CIC, BeCL, and CeSD) in the URLB benchmark. %In skill finetuning, only one skill is used for evaluation. The behavioral diversity is also ignored. It is one-sided to employ skill finetuning alone for evaluation, like recent works.
}
\label{skillfinetuning}

\setlength{\tabcolsep}{3.5mm}{
\begin{tabular}{@{}cc|ccccccc@{}}
\toprule
Domain                     & Task          & DIAYN \cite{diayn}   & SMM \cite{smm}             & APS \cite{aps}   & CIC \cite{cic-nips}            & BeCL \cite{becl}    &   CeSD \cite{cesd}    & ComSD (ours)      \\ \midrule
\multirow{4}{*}{Walker}    & run           & 242±11  & \textbf{430±26}          & 257±27 & \textbf{450±19} & 387±22     & 337±19      & \textbf{447±64}  \\
                           & stand         & 860±26  & 877±34          & 835±54 & \textbf{959±2}  & \textbf{952±2} & \textbf{960±3}  & \textbf{962±9}   \\
                           & flip          & 381±17  & 505±26          & 461±24 & \textbf{631±34} & \textbf{611±18} &541±17 & \textbf{630±41}  \\
                           & walk          & 661±26  & 821±36          & 711±68 & 885±28          & 883±34      &834±34     & \textbf{918±32}  \\ \midrule
\multirow{4}{*}{Quadruped} & run           & 415±28  & 220±37          & 465±37 & 445±36          & 535±13 & \textbf{586±25} & 500±103 \\
                           & stand         & 706±48  & 367±42          & 714±50 & 700±55          & 875±33 & \textbf{919±11}          & 824±86  \\
                           & walk          & 406±64  & 184±26          & 602±86 & 621±69          & 743±68   & \textbf{889±23}       & 735±140 \\
                           & jump          & 578±46  & 298±39          & 538±42 & 565±44          & 727±15 &\textbf{755±14}          & 686±66  \\ \midrule 
\multirow{4}{*}{Hopper}    & hop           & 3±4     & 5±7             &     1±1   & \textbf{59±60}  & 5±7       &12±18       & \textbf{40±35}            \\
                           & flip          & 7±8     & 29±16           &    3±4    & \textbf{96±64}  & 13±15     & 48±49       & 61±47            \\
                           & hop backward  & 9±28    & 29±57           &     2±0   & \textbf{172±64} & 40±72     &117±124       & 92±105           \\
                           & flip backward & 2±1     & 19±34           &    10±23    & \textbf{154±70} & 22±36      &74±71      & 59±63            \\ \midrule
\multirow{4}{*}{Cheetah}   & run           & 151±72  & \textbf{523±17} &    381±41    & 483±32          & 380±35      &375±33     & 432±37           \\
                           & flip          & 615±78  & \textbf{711±6}  &    648±82    & \textbf{730±13} & 701±30   &632±80        & 660±52           \\
                           & run backward  & 368±15  & \textbf{477±19} &     392±41   & 452±11          & 400±20     &387±23      & \textbf{458±9}   \\
                           & flip backward & 477±108 & \textbf{679±7}  &     518±103   & \textbf{678±93} & 643±102   &474±33       & \textbf{685±55}  \\ \bottomrule
\end{tabular}
}

\end{table*}

\subsection{Skill Combination} In skill combination evaluation \cite{diayn}, another meta-controller $\pi'(z|s)$ which selects the discovered skills automatically to achieve downstream tasks, is trained. The discovered skill-conditioned agent $\pi(a|s,z)$ is frozen, and $\pi'(z|s)$ is trained by RL with extrinsic reward (concrete process is provided in Algorithm \ref{algorithm-combination}). As it doesn't allow for adjustments to learned skills with real reward, skill combination is very challenging for unsupervised skill discovery methods, requiring both state exploration and skill diversity. It is one of the most comprehensive skill quality evaluations. For each method, 2M environment steps RL are allowed to train their meta-controllers. All the hyper-parameters are shared across all the methods for a fair comparison, which we detail in Table \ref{hyper}. The results of ComSD are obtained over six different random seeds. 

The training curves for 16 skill combination downstream tasks are shown in Fig. \ref{skillcombination}, where the solid line represents the mean and the shade denotes the standard deviation. DIAYN, SMM, and BeCL cannot obtain effective policies in most downstream tasks (e.g., they all fail in Hopper). CeSD achieves considerable results on Quadruped tasks but fails in the others. These phenomena fully confirm the difficulty of this benchmark for skill discovery methods. In this situation, ComSD enables effective policy learning and outperforms all 6 baselines across 15/16 downstream tasks, demonstrating that the robot behaviors discovered by ComSD are of much higher quality (both state exploration and skill diversity) than all other methods. CIC and APS are the two most competitive baselines. Both of them and our ComSD employ the decomposition Eq. (\ref{eq2}), which indicates the importance of explicitly maximizing exploration. An interesting phenomenon is that CIC always gets a higher initial score than APS and ComSD but has no upward trend, which coincides with the fact that CIC can only produce exploratory but homogeneous behaviors (skill visualization in Section V-J). The results will be similar no matter how these homogeneous skills are combined by the meta-controller. By contrast, APS and ComSD explicitly maximize the conditioned entropy through diversity reward, greatly improving the diversity of learned skill sets and achieving higher final scores than CIC. Compared with APS, ComSD performs better due to the following two reasons: First, the contrastive learning employed by ComSD provides stronger incentives for exploration and exploitation in challenging robot domains. Second, the proposed SMW enables ComSD to learn diverse behaviors at different activity levels, thus gaining better adaptation to both static and dynamic downstream tasks.

\subsection{Skill Finetuning} 
In skill finetuning evaluation \cite{urlb}, a skill vector $z_i$ is chosen, and the corresponding skill $\pi(\cdot|s,z_i)$ is finetuned with the extrinsic reward (concrete process is provided in Algorithm \ref{algorithm-finetuning}). Since reward-specific skill adjustments are allowed, this adaptation is much easier than skill combination. 100k environment steps with extrinsic reward are allowed for each method, where one skill is selected in the first 4k steps and then finetuned in another 96k steps. For skill selection, previous works employ random sampling~\cite{becl}, fixed choice~\cite{cic-nips}, or reward-based choice~\cite{aps}. We follow the official implementation of ~\cite{cic-nips}, employing a fixed mean value (0.5) for all dimensions of the skill vector except the first. We set the first dimension to $0$, because when the first skill dimension (flag variable) is set to 0 (lower than $f_{low}$), the skills have high exploration (see Section IV-C), achieving better results. All the hyper-parameters are shared across different methods for a fair comparison. We show the detailed experimental settings in Table \ref{hyper}. The results of ComSD are obtained over ten different seeds.

\begin{algorithm}[t]
\caption{Pseudo-code of skill finetuning evaluation.}

%\textsc{\#\#\#Adaption evaluation of discovered skills by skill finetuning}

\textbf{Require:} Reward-specific environment $E_s$, the pre-trained skill-conditioned policy $\pi(a|s,z)$, environment adaption steps $I_a$, the RL batch size $I_b$, and skill choice steps $I_c$.

\textbf{Initialize:} The critic $Q(a,concat(s,z))$ and replay buffer $D$.

\begin{algorithmic}[1]
\STATE \hspace{0cm}Choose a skill vector $z_i$ in $I_c$ steps and save the corresponding $I_c$ extrinsic-reward transitions into $D$.
\STATE \hspace{0cm}Freeze the chosen skill vector $z_i$.
\STATE \hspace{0cm}\textbf{for} $t = 1,...,I_a-I_c $ \textbf{do}
%\STATE \hspace{0.5cm}Sample a skill vector from uniform distribution $z_t \sim p(z)$.
%\STATE \hspace{0.5cm}Obtain current skill vector $z_{t} \sim \pi(\cdot|s_t)$ based on current observation $s_t$. 
\STATE \hspace{0.5cm}Obtain action $a_{t} \sim \pi(a|s_t,z_i)$. $s_t$ is the current state.
\STATE \hspace{0.5cm}Interact with reward-specific environment $E_s$ with $a_t$ to get next observation $s_{t+1}$ and the extrinsic reward $r^{extr}$.
\STATE \hspace{0.5cm}Add the transition $(s_t,z_i,a_t,r^{extr},s_{t+1})$ into $D$.
\STATE \hspace{0.5cm}Sample $I_b$ transition batch from $D$.

\STATE \hspace{0.5cm}Use DDPG to update $\pi(a|s_t,z_i)$ and $Q(a,concat(s,z))$ over $I_b$ transitions.

\STATE \hspace{0cm}\textbf{end for}
\end{algorithmic}

\textbf{Output:} The performance of $\pi(a|s,z_i)$ serves as the skill finetuning evaluation result.

\label{algorithm-finetuning}
\end{algorithm}

The numerical results on 16 downstream tasks are shown in Table \ref{skillfinetuning}, where the first value represents the mean and the second value denotes the standard deviation. CeSD, CIC, and BeCL are current state-of-the-art methods in URLB skill finetuning. Compared with CIC and BeCL, ComSD performs comparably or better on 11 and 12 downstream tasks, respectively. It also obtains competitive or higher scores than CeSD on 11 downstream tasks. These demonstrate that ComSD achieves competitive performance with current state-of-the-art methods. Actually, the score of the initial few steps has little to do with the final score in state-based locomotion. It's hard to find a proper skill from a large skill set within only 4K steps. This explains why CIC, BeCL, CeSD, and our ComSD use random or fixed skill choices to achieve the advanced performance. Meanwhile, it means this evaluation does not adequately judge the skill diversity (only one skill is employed for evaluation), and is even less favorable to a more diverse skill set, because higher skill diversity causes higher skill choice difficulties. In this situation, ComSD discovers the most diverse robot behaviors (see skill combination in Section V-C and diversity analysis in Section V-G\&H) while simultaneously achieving competitive performance on skill finetuning, which further shows its superior behavioral quality.

\begin{figure*}[t]
  \centering
  \setlength{\abovecaptionskip}{0.cm}
  \subfloat{
    \includegraphics[width=0.45\textwidth]{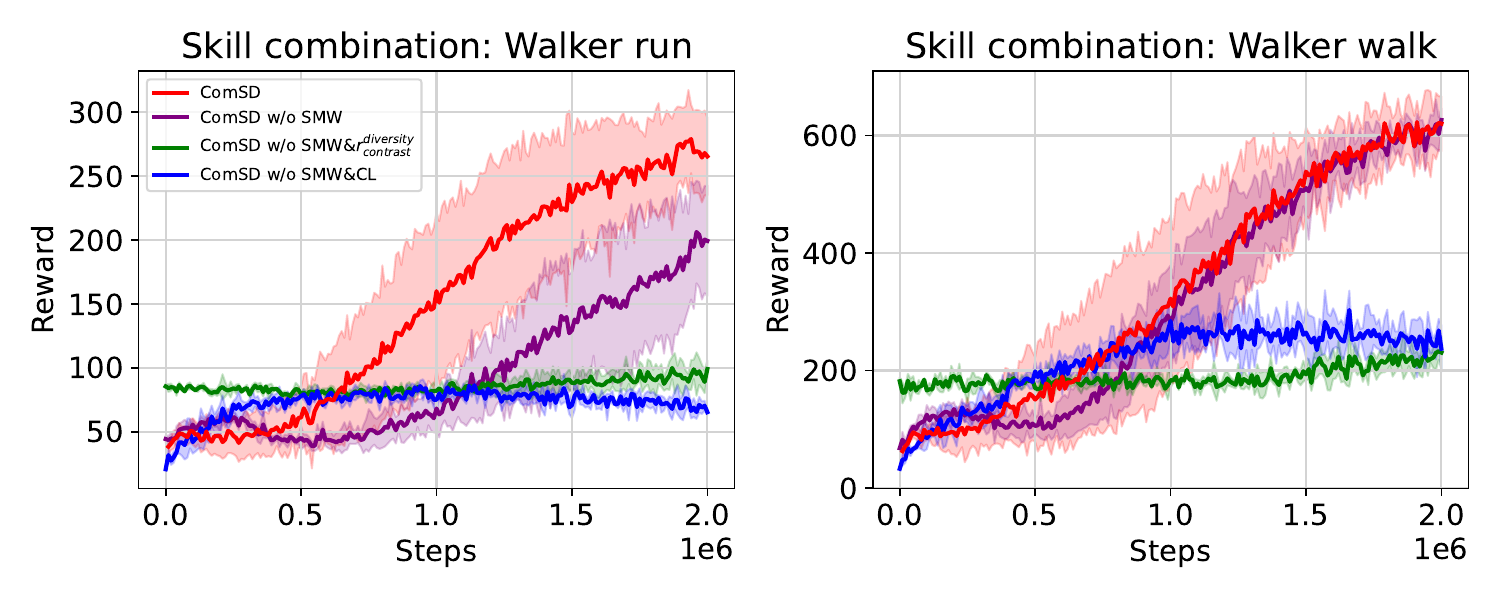}
    %\caption{skill combination evaluation}
    \label{fig:subfig1}
  }
  \subfloat{
    \includegraphics[width=0.45\textwidth]{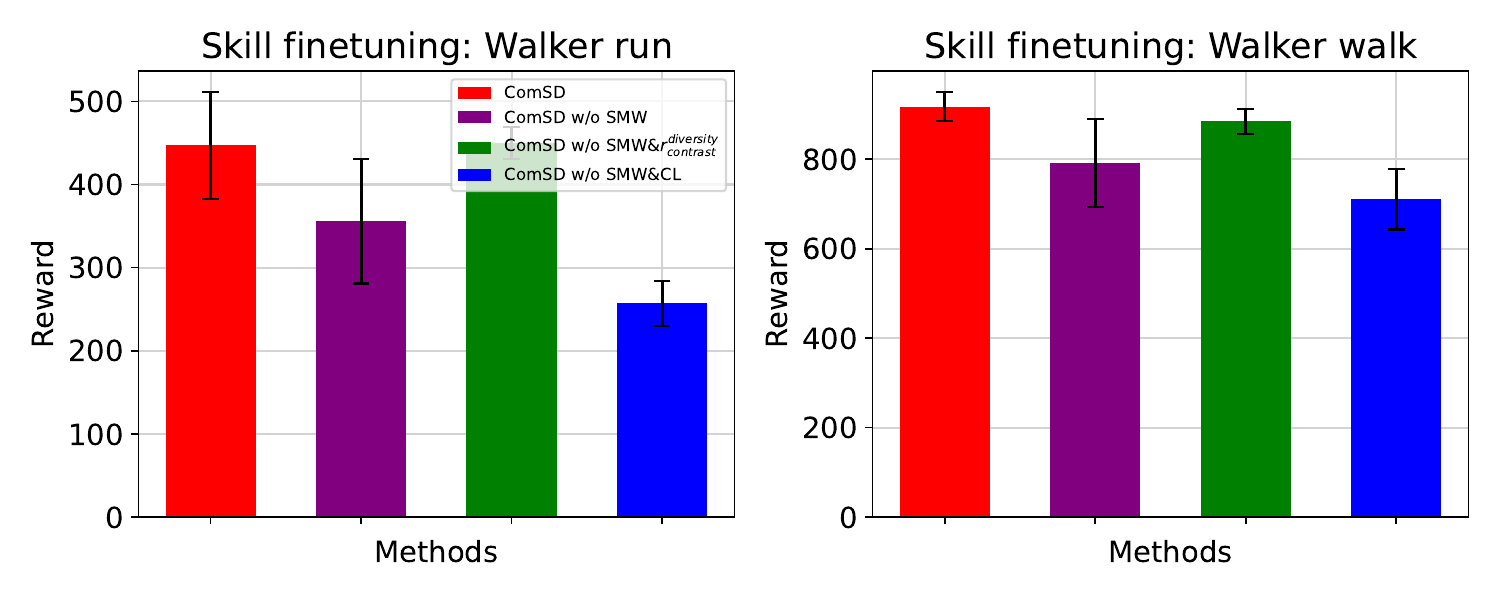}
    %\caption{fast adaption evaluation}
    \label{fig:subfig2}
  }
  \vspace{-2mm}
  \caption{Adaptation ablation experiments on (left two) skill combination tasks and (right two) skill finetuning tasks. Our contrastive diversity reward ($r^{diversity}_{contrast}$) and skill-based dynamic weighting (SMW) are both necessary for ComSD to achieve the advanced results on both kinds of adaptation evaluations. }
  \label{ablation}
  \vspace{-3mm}
\end{figure*}

\begin{figure*}[t]
  \centering
  \subfloat{
    \includegraphics[width=0.46\textwidth]{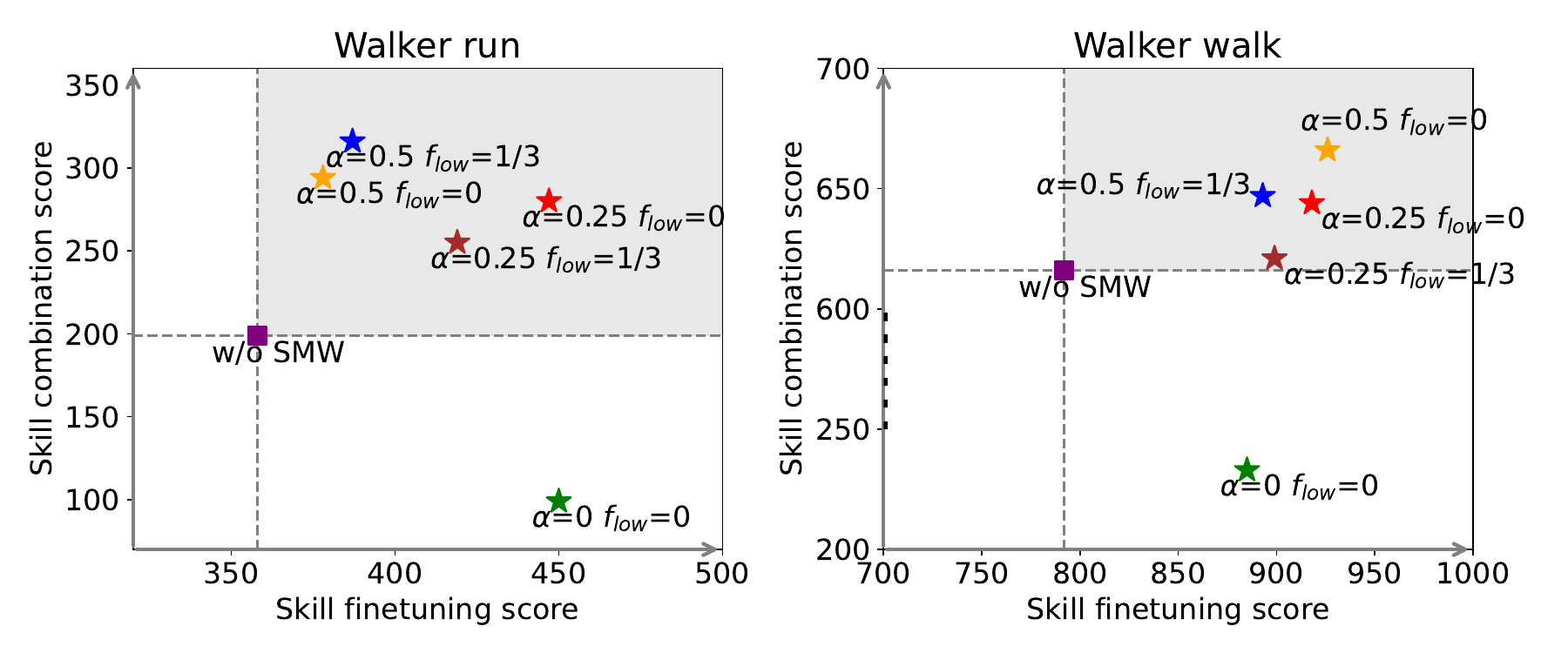}
  }
  \subfloat{
    \includegraphics[width=0.46\textwidth]{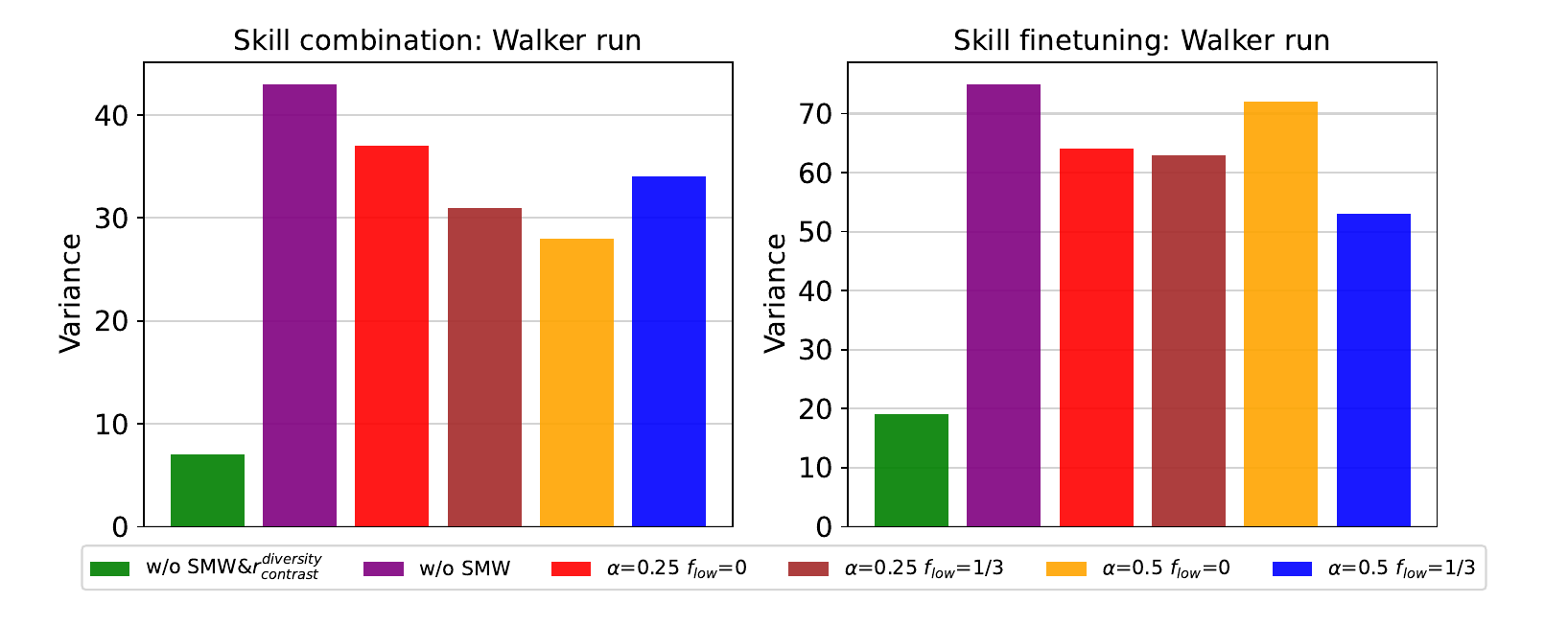}
  }
  \vspace{-1mm}
  \caption{Hyper-parameter sensitivity analysis of SMW. $\alpha$ determines the maximum value of the dynamic weight, and $f$ determines the variation range of the dynamic weight. When $f_{low}=0$, $f_{high}$ is correspondingly set to $1$, while when $f_{low}=1/3$, $f_{high}$ is correspondingly set to $2/3$. We omit $f_{high}$ in the figure for simplicity.  \textbf{Left two:} The performance comparison between different SMW hyper-parameters. SMW is effective across multiple hyper-parameter settings. \textbf{Right two:} The variance comparison between different SMW hyper-parameters. The diversity reward significantly increases the performance variance, while SMW doesn't apparently affect the variance. }  
  \label{sensitivity}
  \vspace{-3mm}
\end{figure*}

%\section{Experiments: Behavioral Quality Analysis}
\label{section-4-3}

\subsection{Adaptation Ablation}
We conduct adaptation ablation experiments to show the effectiveness of (i) the contrastive diversity reward $r^{diversity}_{contrast}$ and (ii) the skill-based dynamic weighting (SMW) mechanism in Fig. \ref{ablation}. 'ComSD w/o SMW' denotes ablating the SMW, i.e., removing the dynamic weight $\beta(z)$ in Eq. (12). 'ComSD w/o SMW\&$r^{diversity}_{contrast}$' denotes removing the contrastive diversity reward, i.e., only employing the exploration reward $r^{exploration}$. 'ComSD w/o SMW\&CL' denotes that based on 'ComSD w/o SMW', Contrastive Learning (CL) is further removed, which is actually APS \cite{aps}. \. %, where CIC is actually ComSD w/o SMW\&$r^{diversity}_{contrast}$, and APS is actually ComSD w/o contrastive learning\&SMW. 
Compared with no diversity reward ('ComSD w/o SMW\&$r^{diversity}_{contrast}$'), designing a concrete diversity reward ('ComSD w/o SMW\&CL' and 'ComSD w/o SMW') enables better skill diversity, which corresponds to the larger curve slope in skill combination. Moreover, contrastive learning enables better performance on both kinds of adaptations. In practice, a fixed diversity reward coefficient ('ComSD w/o SMW\&CL' and 'ComSD w/o SMW') cannot provide a balance between state exploration and skill diversity. The direct intervention of diversity reward indeed improves diversity but simultaneously reduces exploration (more details in Section V-I), causing huge performance drops in skill finetuning. The proposed SMW enables agents to balance them well, effectively alleviating the state exploration decline and corresponding performance drops in skill finetuning. In summary, both $r^{diversity}_{contrast}$ and SMW are necessary in ComSD.

\subsection{Hyper-parameter Sensitivity} In this section, we provide the hyper-parameter selecting process (i.e., sensitivity analysis) of SMW. We test different $\alpha$ values and different $f$ values (containing $f_{low}$ and $f_{high}$) in SMW. $\alpha$ determines the maximum value of the dynamic weight, and $f$ determines the variation range of the dynamic weight. The performance comparison is shown in Fig. \ref{sensitivity} left. The stars lying in the gray space mean that both the skill finetuning and skill combination performance are improved through SMW. When $\alpha$ is set to 0 (green), the diversity reward $r^{diversity}_{contrast}$ is actually removed, leading to low skill diversity and low skill combination performance. With $\alpha$ increased to 0.25 (red), SMW enables much higher skill quality, leading to better adaptation results across both evaluations. Continuing to increase $\alpha$ (yellow, blue) or adjusting the variation range (brown, blue) does not lead to consistent improvements on different tasks, which makes us choose hyper-parameters corresponding to the red star. In addition, there are four stars in the gray space, which means the proposed SMW is effective across multiple hyper-parameter settings. 

In addition, we also provide the variance comparison between different hyper-parameter settings in ComSD. The results (Fig. \ref{sensitivity} right) demonstrate that the diversity reward significantly increases the performance variance. The higher skill diversity inevitably leads to greater uncertainty in both agent pre-training and skill adaptation (e.g., there are more potential combinations of skills), which is consistent with our intuition. By contrast, SMW is designed to alleviate the exploration hurt brought by the use of our diversity reward. It doesn't apparently affect the variance.

\subsection{State Exploration \&  Behavioral Activity }  In this section, we directly analyze the state exploration of discovered skills, where a metric to evaluate the exploration of one skill is needed. Here we utilize the visited-state coverage of one skill to represent its exploration level. Concretely, we use one skill to sample 1K states in the environment. Then the visited-state coverage can be estimated by the All-K-nearest-neighbor Distance (AKD)~\cite{apt} that computes the mean Euclidean distance between each state and its k nearest neighbors. AKD is proven to be a good estimator~\cite{apt,proto-rl} of state coverage, thus it can represent the exploration level of one skill. For a robot behavior, more visited states usually mean it is more active. To this end, AKD can also partially represent the activity level of one robot behavior.

\begin{figure*}[t]
  \centering

   \includegraphics[width=0.95\textwidth]{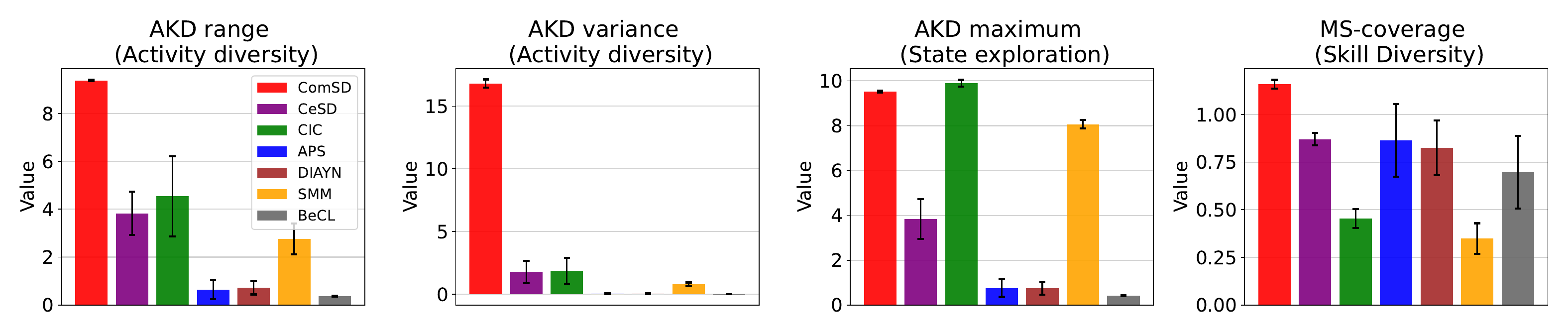}
   \vspace{-2mm}

  \caption{ \textbf{Left three}: Behavioral activity and state exploration analysis of all methods. AKD can represent the exploratory level and activity level of one robot behavior. ComSD and CIC lead on AKD maximum, which means they discover the most exploratory skills, achieving higher state exploration than other baselines. The much higher scores on AKD range and variance mean ComSD generates robot behaviors at different activity levels (i.e., higher activity diversity), including dynamic movements and static postures, whereas other methods cannot. \textbf{Right one}: Skill diversity analysis of all methods. ComSD discovers more diverse robot behaviors. }
  \label{skillanalysis}
  \vspace{-3mm}
\end{figure*}

\begin{figure}[t]
  \centering
  
  \setlength{\abovecaptionskip}{0.cm}
  \vspace{-3mm}
  \subfloat{
    \includegraphics[width=0.24\textwidth]{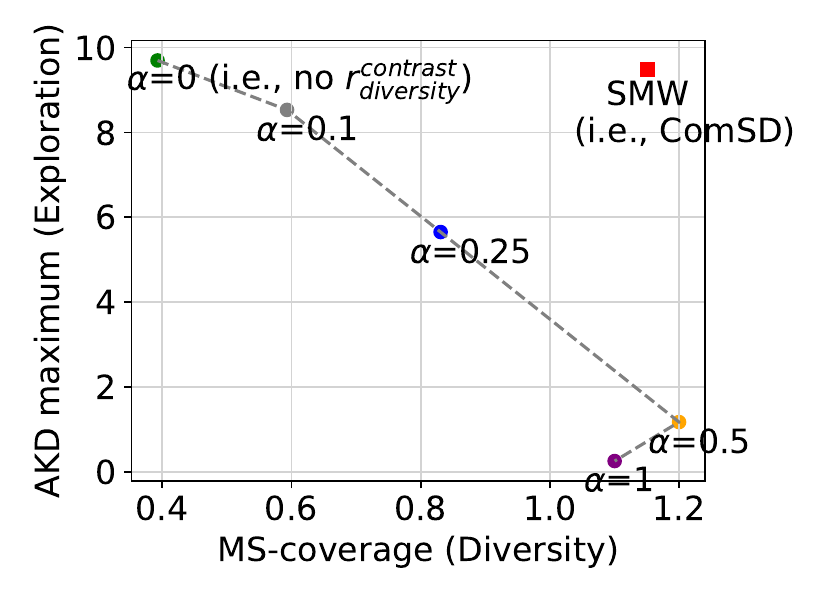}
    %\caption{skill combination evaluation}
    \label{fig:subfig1}
  }
  \hspace{-0.03\textwidth}
  \subfloat{
    \includegraphics[width=0.24\textwidth]{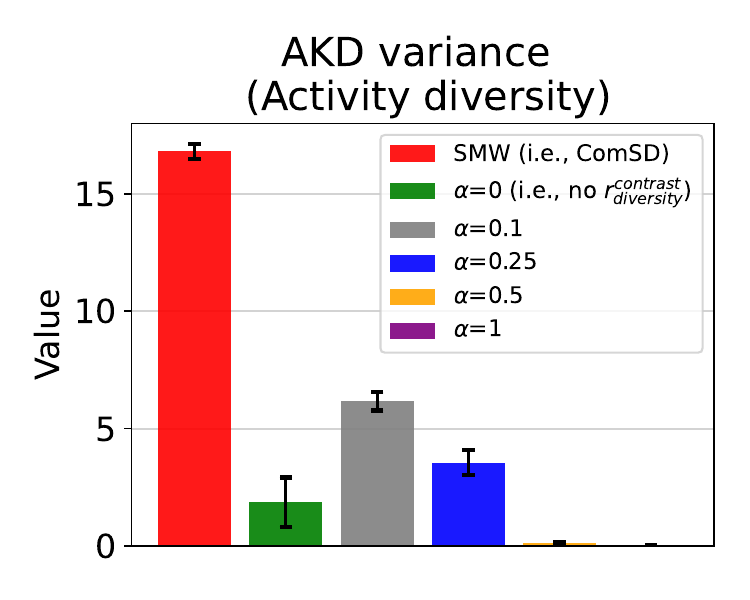}
    %\caption{fast adaption evaluation}
    \label{fig:subfig2}
  }
  \caption{ SMW motivation \& Behavioral quality ablation. We ablate SMW, setting different fixed coefficients $\alpha$ of $r^{diversity}_{contrast}$ to show the effect on behavioral quality. \textbf{Left}: The conflict between state exploration and skill diversity in challenging robot domains. Increasing the fixed weight $\alpha$ of diversity reward can surely improve skill diversity, but it will simultaneously reduce state exploration. The proposed SMW well balances them, effectively alleviating the issue above. \textbf{Right}: SMW enables agents to discover robot behaviors at different activity levels, further improving skill diversity and behavioral quality. }
  \label{motivation}
\end{figure}

For each method, we use grid sweep ~\cite{cic-nips} to select eleven Walker skills and calculate the AKD value (representing exploration level and activity level) of each skill. Then, we calculate the variance, range, and maximum of AKD over all selected skills for each method. The results are shown in Fig. \ref{skillanalysis}, where the error bar denotes the standard deviation. ComSD and CIC lead on AKD maximum, which means they discover the most exploratory and complex behaviors, accessing more far-reaching states and gaining higher state exploration than other baselines. By contrast, other methods like APS can only produce lazy behaviors (skill visualization in Section V-J), which corresponds to its low score on AKD maximum metric. In addition, ComSD exhibits huge superiority on AKD range and AKD variance, which means it discovers behaviors at different activity levels (i.e., high activity diversity), including both dynamic movement and static postures. This explains why ComSD achieves state-of-the-art scores across both dynamic downstream tasks (e.g., Walker run) and static downstream tasks (e.g., Walker stand). %This is mainly attributed to multiple optimization targets brought by multi-objective intrinsic reward. 

%We strongly refer readers to Appendix \ref{AppendixD} for additional visualization and analysis.

\subsection{Skill Diversity}
In this section, we try to analyze the skill diversity of all methods. Intuitively, a more diverse skill set should cover a wider skill space. To this end, we first need to represent each skill, which is achieved by utilizing its visited states. Concretely, we use one skill to sample 1K states in the environment and obtain the Mean State (MS) vector of all 1K states. This MS vector can partially represent one skill. Note that the AKD in Section V-G is used to represent the exploration and activity level of one skill, while the MS vector here directly represents the skill itself. 

For each method, we use grid sweep ~\cite{cic-nips} to select eleven Walker skills, and then calculate the skill coverage (MS-coverage) by computing the mean k-nearest-neighbor Euclidean distance over selected skills (represented by MS vector)~\cite{apt}. MS-coverage can represent skill coverage, i.e., skill diversity, to some extent. The results in Fig. \ref{skillanalysis} demonstrate that ComSD has huge advantages on MS-coverage, which means the highest skill diversity. ComSD also discovers behaviors at diverse exploration levels (higher AKD range and higher AKD variance), which have been analyzed in Section V-G. In conclusion, ComSD indeed produces more diverse skill sets than other baselines.

%Obviously, it's hard to represent an exploratory skill by MS, which puts highly exploratory methods (ComSD and CIC) at a disadvantage. In this case, CIC fails on these metrics, while ComSD still obtains state-of-the-art results. which demonstrates that the behavior set discovered by ComSD is of much higher diversity and coverage than other baselines.

\begin{figure}[t]
  \centering
  \subfloat{
    \includegraphics[width=0.45\textwidth]{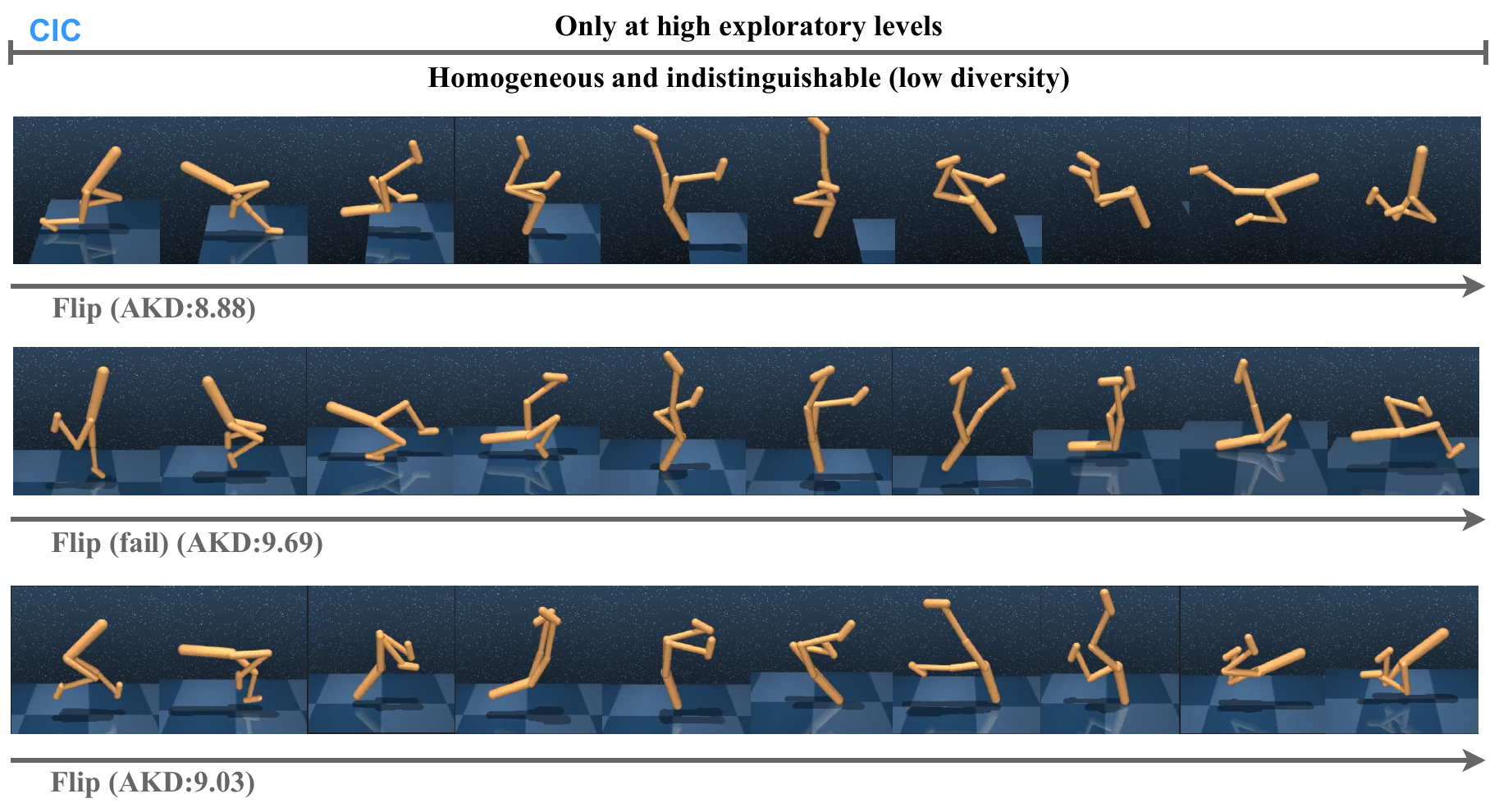}
    %\caption{skill combination evaluation}
    \label{fig:subfig1}
  }
  \\
  \subfloat{
    \includegraphics[width=0.45\textwidth]{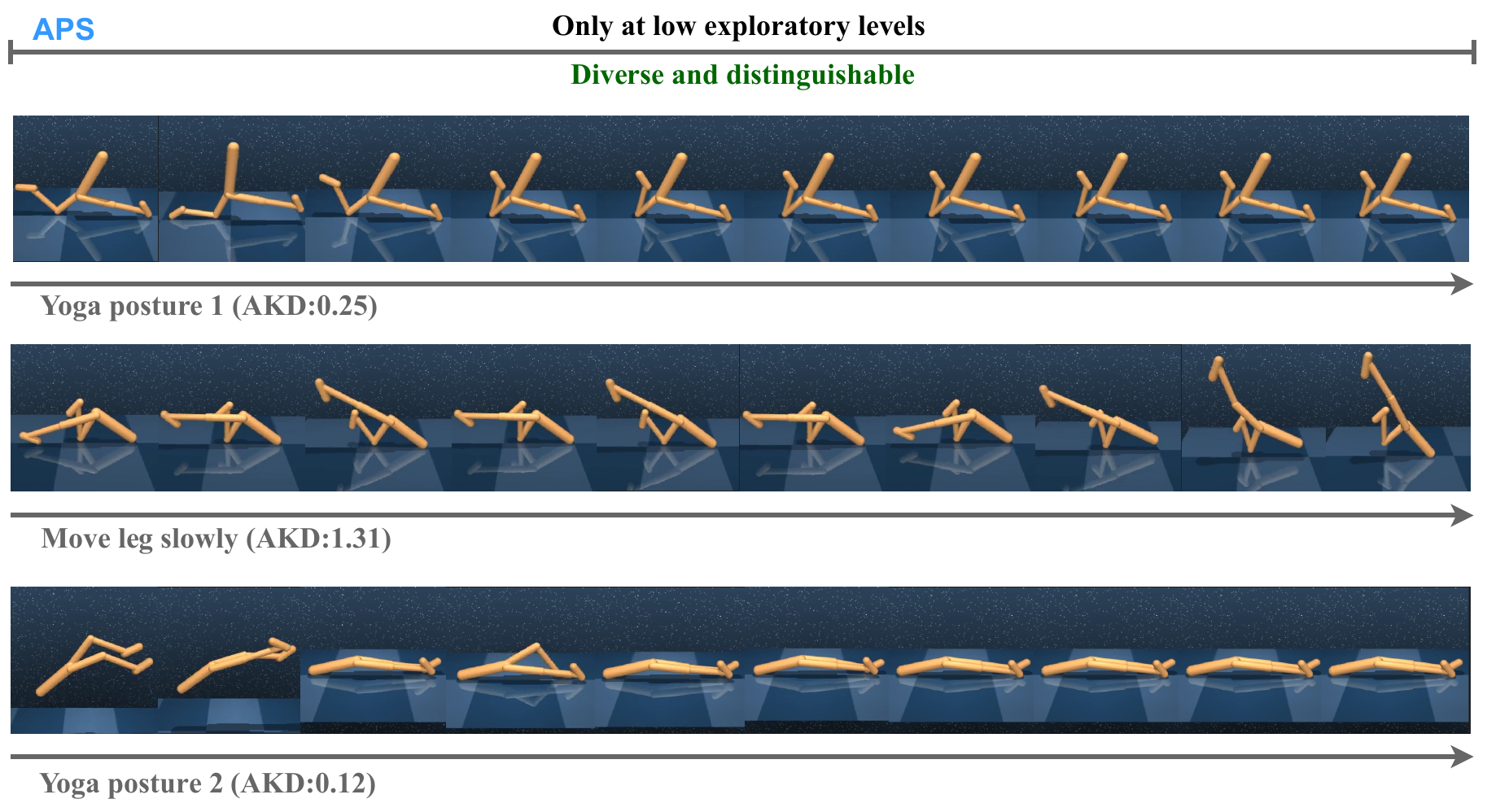}
    %\caption{fast adaption evaluation}
    \label{fig:subfig2}
  }
  \caption{Visualization for representative robot behaviors discovered by CIC and APS. AKD is a state entropy estimator used to evaluate state exploration (activity) of one skill, which we define in Section V-G. Skill AKD ranges from 8 to 10 in CIC and 0 to 2 in APS, which means they can't generate behaviors at different activity levels. In addition, CIC's behaviors are homogeneous while APS's behaviors are all lazy. }
  \label{cicaps}
\end{figure}

\begin{figure}[t]
  \centering
  \subfloat{
    \includegraphics[width=0.45\textwidth]{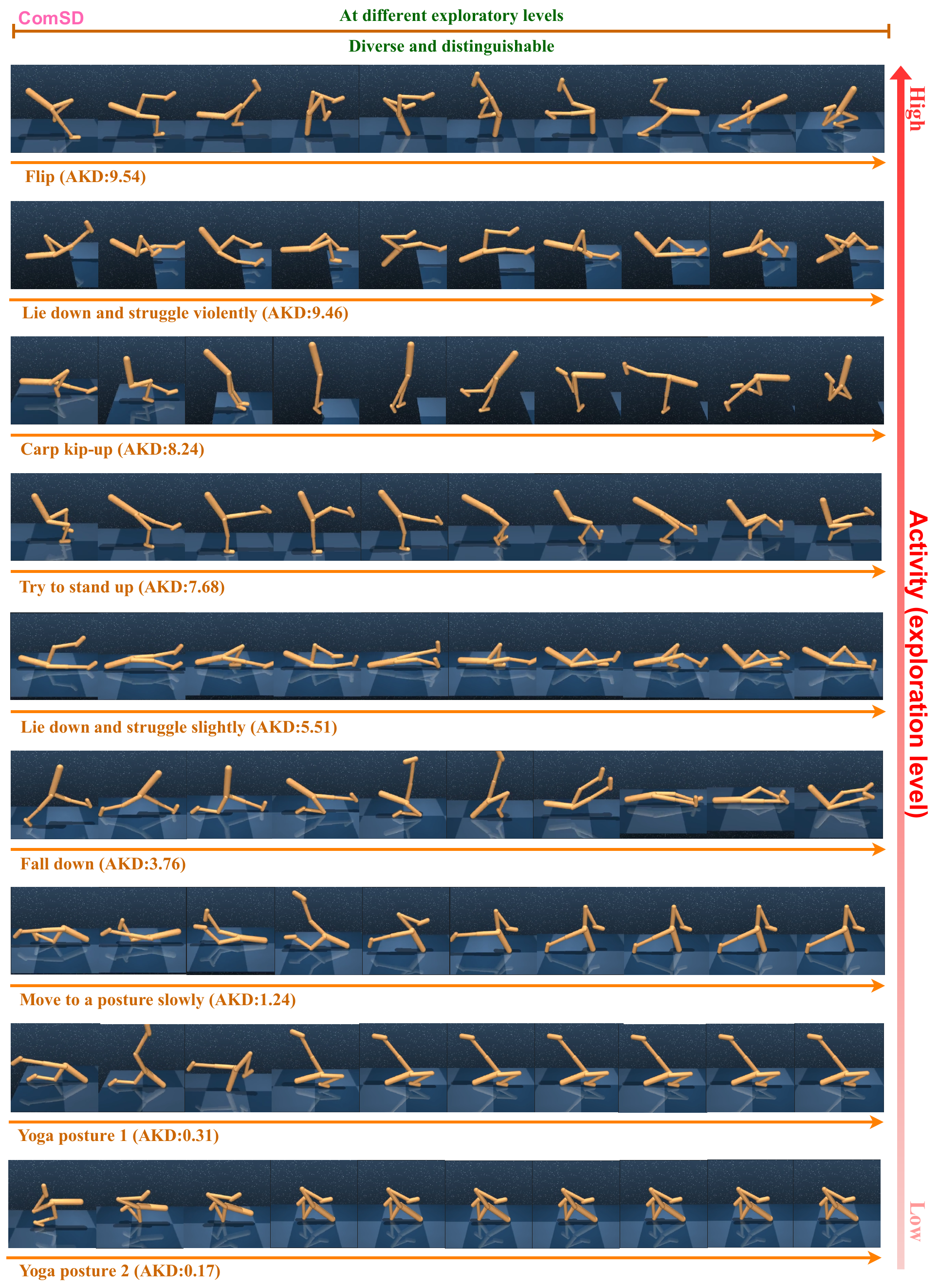}
  }
  \caption{Visualization for representative robot behaviors discovered by ComSD. AKD is a state entropy estimator used to evaluate state exploration (activity) of one skill, which we define in Section V-G. Skill AKD ranges from 0 to 10 in ComSD. The results verify that our ComSD can produce diverse behaviors at different activity levels, which recent advanced methods, such as CIC and APS, cannot.}
  \label{comsdvisual}
\end{figure}

\subsection{Diversity-exploration Conflict \& Quality Ablation Study \& The motivation of SMW} We ablate SMW, setting different fixed coefficients $\alpha$ of $r^{diversity}_{contrast}$ to show the effect on behavioral quality. As shown in Fig. \ref{motivation} left, there exists a serious conflict between state exploration and skill diversity when using a fixed coefficient (i.e., without SMW). Concretely, $r^{diversity}_{contrast}$ simultaneously improves skill diversity and hurts state exploration, and so do other diversity incentives in other methods (e.g., APS). Any fixed weight $\alpha$ cannot well balance both of them. This motivates us to design SMW that links the weight to the changing skill vectors, achieving dynamic weighting, as mentioned in Section IV-C. In addition, SMW further enables behavior discovery at different activity levels (Fig. \ref{motivation} right), which a fixed coefficient also cannot support.

In Fig. \ref{skillanalysis}, we observe that CIC and SMM lead on AKD maximum (state exploration) but obtain the lowest scores on MS-coverage (skill diversity). The rest of the baselines with higher MS-coverage (skill diversity) all suffer from lazy exploration. These demonstrate that the conflict between exploration and diversity also exists in all the baseline methods.

\subsection{Skill Visualization: What Robot Skills Do ComSD and Competitive Baselines Discover?} We provide the skill visualization and corresponding skill exploration level (AKD) of the two most competitive baselines, APS~\cite{aps} and CIC~\cite{cic-nips}, in Fig. \ref{cicaps}. The visualization and skill AKD of our ComSD are shown in Fig. \ref{comsdvisual}. AKD represents the state exploration (activity) of one skill, which we define in Section V-G.

CIC is able to produce continuous and dynamic movements of high exploration, but it can't generate behaviors at other activity levels (AKD range of CIC's skills is 8-10). In addition, CIC suffers from insufficient skill diversity, i.e., the generated skills are indistinguishable and homogeneous. CIC's skills all tend to achieve dynamic flipping, which is consistent with the good initial score on 'Walker flip' in skill combination (Fig. \ref{skillcombination}). However, the low skill diversity makes it difficult for meta-controllers to find a better combination for competitive final scores. APS can generate non-homogeneous robot skills, but it suffers from lazy state exploration and also can't generate behaviors at different activity levels (AKD range of APS's skills is 0-2). Compared with CIC, the higher skill diversity allows the meta-controller to partially complete downstream tasks in skill combination, which coincides with the upward trend of APS training curves (Fig. \ref{skillcombination}). However, poor exploration causes poor performance in skill finetuning, simultaneously. In summary, previous advanced methods can't provide a good balance between state exploration and skill diversity, thus failing to exhibit competitive results across different kinds of adaptation evaluations. 

By contrast, our ComSD can produce diverse behaviors at different exploratory levels (AKD range of ComSD's skills is 0-10), including flipping, lying down, struggling at different speeds, various postures, and so on. ComSD's huge superiority in behavioral quality explains why it achieves state-of-the-art adaptation performance across both kinds of downstream tasks, while other methods cannot. We further provide the skill videos in \url{https://github.com/liuxin0824/ComSD}.

\begin{figure*}[t]
  \centering
  \subfloat{
    \includegraphics[width=0.93\textwidth]{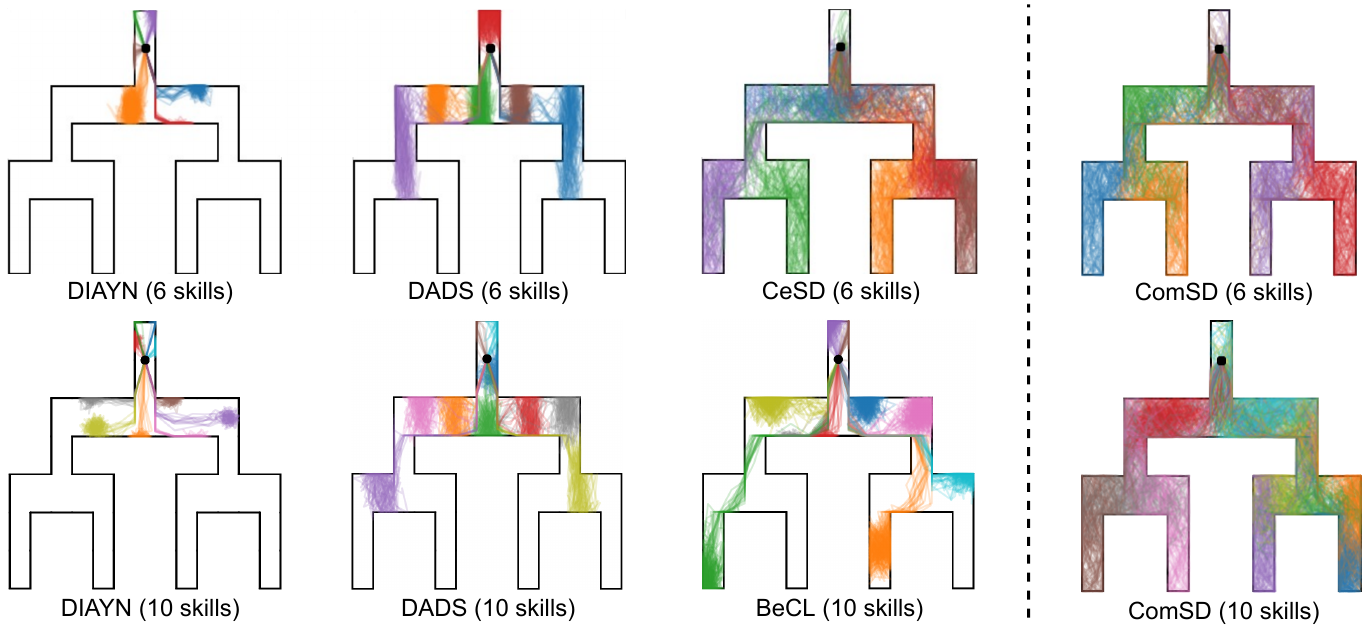}}
   \vspace{-3mm} 
  \caption{The exploration skill discovery on the 2D maze. In both 6-skill and 10-skill settings, ComSD can cover all the maze areas, reach the deepest leaves, and obtain the distinguishable exploration skills. The results mean that our ComSD achieves state-of-the-art exploration skill discovery performance on the challenging 2D tree-like maze.}
  \label{treemaze}
  \vspace{-3mm} 
\end{figure*}

\subsection{Skill Visualization: 2D Maze Exploration Skill Discovery} In this section, we further evaluate our ComSD on 2D maze exploration \cite{map-exploration-env}. We compare ComSD with state-of-the-art maze exploration methods: BeCL \cite{becl} and CeSD \cite{cesd}. Following these methods, we choose the most challenging tree-like maze, which consists of two branches and four leaves. In this maze, BeCL trains 10 skills simultaneously, while CeSD only provides the result of 6 skills in their papers. To this end, we conducted experiments in both settings. Due to the discrete skill space, SMW maps the diversity reward weight of all skills to the range of 0.5-1 at the same interval to achieve dynamic weighting in ComSD. DIAYN \cite{diayn} and DADS \cite{dads} are also included in comparison.

The results are shown in Fig. \ref{treemaze}. In 6-skill experiments, both CeSD and our ComSD can cover all the maze areas, reach the deepest leaves, and lean distinguishable skills. The other baselines, DIAYN and DADS, can't achieve this performance. Compared with CeSD, ComSD's 6 skills respectively reach four leaves (blue, yellow, purple, and red) and two branches (green, brown), enabling a more even skill distribution. In the 10-skill experiment, ComSD can fully explore all the branches and leaves while all the other baselines fail to do that. The learned 10 skills can be easily discerned by their colors. In summary, ComSD can achieve the leading skill discovery results on the challenging tree-like maze exploration. ComSD is also the only method that can well balance state exploration and skill diversity in both maze and robot environments.

\subsection{Additional Comparison with Distance-Maximizing Methods}

In this section, we make further comparisons with three advanced distance-maximizing skill discovery methods: LSD\cite{lsd}, CSD\cite{csd}, and METRA \cite{metra}, on different kinds of downstream tasks. These methods are not evaluated on the URLB benchmark, so we employ the re-implemented results provided by recent advanced works \cite{cesd}. Results in Table \ref{lsdcsdmetra} demonstrate that our ComSD achieves leading adaptation results across eight tasks, while the distance-maximizing methods don't perform very well. This is because these methods are mainly designed to discover advanced goal-reaching skills (e.g., reaching far goals in different directions), while the URLB benchmark consists of different kinds of downstream tasks, requiring the agent to discover different body movements (e.g., jump, flip, and stand). Their discovered skills can't well match the target tasks defined in URLB without finetuning. Further, these goal-reaching skills usually don't focus on exploring complex body movements, which also leads to inefficient exploration and understanding of the high-reward space during the extrinsic adaptation. For example, the quadruped of METRA learns to roll gently on the ground in different directions, which makes it not easy to explore the rewards associated with jumping or walking. In addition, the greater difficulty of the URLB environments (DMC) compared to OpenAI Gym (according to \cite{cic-nips}), as well as the lack of official hyper-parameters during re-implementation \cite{cesd}, could both potentially be factors affecting the performance. In contrast, ComSD aims to discover different complex robot body movements, which makes it more suitable for the URLB.

\begin{table}[t]
\centering
\renewcommand\arraystretch{1.1}
\vspace{-4mm} 
\caption{The skill finetuning comparison with advanced distance-maximizing skill discovery methods. ComSD enables better adaptation performance across eight tasks.} 
\setlength{\tabcolsep}{2mm}{

\begin{tabular}{@{}ccccc@{}}
\toprule
Task            & LSD \cite{lsd}    & CSD \cite{csd}            & METRA  \cite{metra}         & ComSD (ours)     \\ \midrule
Walker run      & 130±22  & \textbf{457±50} & 361±45          & \textbf{447±64}  \\
Walker stand    & 837±3   & \textbf{942±8}  & \textbf{943±13} & \textbf{962±9}   \\
Walker flip     & 223±6   & 602±11          & 589±75          & \textbf{630±41}  \\
Walker walk     & 323±75  & 802±85          & 850±63          & \textbf{918±32}  \\
Quadruped run   & 270±55  & 329±62          & 196±34          & \textbf{500±103} \\
Quadruped stand & 426±131 & 425±120         & 324±173         & \textbf{824±86}  \\
Quadruped walk  & 256±83  & 353±142         & 190±44          & \textbf{735±140} \\
Quadruped jump  & 247±54  & 520±80          & 224±17          & \textbf{686±66}  \\ \bottomrule
\end{tabular}}

\label{lsdcsdmetra}
\vspace{-2mm} 

\end{table}

\subsection{Transferred to Pixel-based Tasks} 
According to previous works \cite{urlb,cesd}, current advanced unsupervised skill discovery methods perform poorly when directly transferred into pixel-based URLB with model-free backbone RL. We attribute this to the agents' poor understanding of high-dimensional visual inputs. To this end, we employ an extra contrastive target between neighboring states ($s^{t-1},s^{t}$) along with the original contrastive target between state pairs and skill vectors ($\tau,z$), to enhance image understanding of ComSD (the states denote image inputs here). This target can be directly obtained in ComSD because the state pair actually consists of two neighboring states, i.e., $\tau = s^{t-1},s^{t}$. The extra contrastive target is computed by replacing ($\tau,z$) in Eq. (7) with ($s^{t-1},s^{t}$). The results are shown in Table \ref{pixel}, where all skill discovery methods employ the model-free DrQv2 \cite{drq-v2} as their backbone RL algorithm. The Backbone in Table \ref{pixel} denotes the end-to-end results of DrQv2 without skill pre-training to reflect whether the unsupervised skill pre-training of each method is helpful. Most current advanced methods, such as CIC \cite{cic-nips} and CeSD \cite{cesd}, can't enable effective pre-training with model-free backbone RL on pixel-based tasks. The pre-trained agents may get stuck in a certain posture and can't achieve effective downstream RL in some tasks. All of these phenomena are consistent with observations and conclusions in previous works \cite{cesd,becl,urlb}. SMM \cite{smm} achieves the highest results in state-based Cheetah, but it also inhibits its downstream RL when directly transferred to pixel settings. By contrast, ComSD can produce dynamic movements and effectively improve the downstream RL performance across different tasks. Although ComSD achieves improvement upon its model-free backbone RL in pixel-based settings, it still performs worse than ComSD in state-based settings (see Table \ref{skillfinetuning} for state-based results). How to reduce this gap is another big and challenging topic, deserving further specialized studies.

\begin{table}[t]
\centering
\renewcommand\arraystretch{1.1}
\caption{The skill finetuning results on pixel-based Cheetah tasks.  ComSD can improve the performance of its model-free backbone RL in pixel-based settings. } 
\label{pixel}
\setlength{\tabcolsep}{0.8mm}{

\begin{tabular}{cccccc}
\toprule
Task (pixel)          & \begin{tabular}[c]{@{}c@{}}Backbone\\ \cite{drq-v2}\end{tabular} & \begin{tabular}[c]{@{}c@{}}CIC\\ \cite{cic-nips}\end{tabular} & \begin{tabular}[c]{@{}c@{}}CeSD\\ \cite{cesd}\end{tabular} & \begin{tabular}[c]{@{}c@{}}SMM\\ \cite{smm}\end{tabular} & \begin{tabular}[c]{@{}c@{}}ComSD\\ (ours)\end{tabular}        \\ \midrule
Cheetah run           & 226±35                                                    & 7±3                                                    & 9±3                                                 & 188±15                                            & \textbf{315±16} \\
Cheetah flip          & 337±270                                                   & 194±32                                                 & 190±11                                              & 273±128                                           & \textbf{656±17} \\
Cheetah run backward  & 332±12                                                    & 16±2                                                   & 20±5                                                & 215±6                                             & \textbf{409±39} \\
Cheetah flip backward & 239±270                                                   & 166±35                                                 & 189±7                                               & 173±12                                            & \textbf{629±10} \\ \bottomrule
\end{tabular}}

%\vspace{-3mm}

\end{table}

\section{Conclusion \& Limitation}
In this paper, we propose ComSD, a novel unsupervised skill discovery approach, to balance state exploration and skill diversity in challenging domains where potential skills are rich and hard to distinguish. This is achieved through a novel diversity incentive based on contrastive learning, and a novel dynamic weighting mechanism based on skill vectors themselves. We conduct extensive experiments and detailed analysis on challenging benchmarks, verifying the superiority of ComSD in both robot behavior discovery and 2D maze exploration skill discovery over advanced baselines. We hope (i) our ComSD can inspire more attention to the exploration-diversity trade-off in the unsupervised RL community, and (ii) our comprehensive experiments can inspire unsupervised skill evaluation for future works. There are also some limitations. Although ComSD can improve the downstream RL in pixel-based settings, it still performs worse than that in state-based settings. How to well utilize skill discovery methods in pixel settings is another big and challenging topic, which is not adequately discussed in this paper but deserves further specialized studies. In addition, SMW introduces the idea of dynamic weights, but still requires adjusting the weight range for different domains with different agent dynamics. Learning this range automatically may be a better solution, which we leave to future work.

% \usepackage{booktabs}

\begin{table}[h]
%\scriptsize
\caption{Detailed Hyper-parameter settings of ComSD. ($f_{low},f_{high}$) is set to ($1/3,2/3$) for Hopper \& Cheetah. %$\alpha$ is set to $5e-3$ for pixel-based Cheetan. 
In maze exploration, the skills are discrete. SMW maps the diversity reward weight of all skills to the range of 0.5-1 at the same interval to achieve dynamic weighting. }
\centering
\renewcommand\arraystretch{1.1}  
\setlength{\tabcolsep}{2mm}{
\begin{tabular}{|lc|}
\hline
Hyper-parameter                      & Setting       \\ \hline
Skill vector dimensions             & $64$ \\
Skill vector space            & $[0,1] $   continuous   \\
Skill update frequency                   & $50$           \\
%Embedding MLP in $f_{\theta_1}(\cdot)$  & ${\rm dim}(s) \rightarrow 1024$  \\ 
%    &  $\rightarrow 1024 \rightarrow 64 $       \\
%Predictor (MLP) in $f_{\theta_1}(\cdot)$  & $ 64\times 2 \rightarrow 1024$ \\
%    &  $ \rightarrow 1024 \rightarrow 64 $       \\
%State encoder activation        & ${\rm ReLU}$           \\
%Skill encoder (MLP) $f_{\theta_2}(\cdot)$  & $64 \rightarrow 1024 $   \\
%& $\rightarrow 1024 \rightarrow 64 $       \\
%Skill encoder activation        & ${\rm ReLU}$           \\
($w_{low}$,$w_{high}$)        & $(0,2)$\\
%$\beta$ lower bound         & $0$\\
($f_{low}$,$f_{high}$)     & $(0,1)$ \\
$\alpha$  for Walker      & $0.25$   \\
$\alpha$   for Quadruped     & $1e-3$           \\
%$f_{high}$ for hopper \& cheetah & $2/3$   \\
%$f_{low}$   for hopper \& cheetah  & $1/3$ \\
$\alpha$  for Hopper       & $1.25$            \\
$\alpha$  for Cheetah     & $1$           \\

RL backbone algorithm & DDPG \\
Pre-training steps & $2e+6$ \\
RL replay buffer size & $1e+6$ \\
Temperature $T$  & $0.5$  \\
Action repeat & $1$\\
Seed (random) frames & $4000$ \\
Return discount & $0.99$\\
Number of discounted steps & $3$\\
Batch size & $1024$ \\
Optimizer & Adam \\
Learning rate & $1e-4$\\
%Actor network (MLP) & ${\rm dim}(s)+64 \rightarrow 1024 $  \\
%&   $ \rightarrow 1024 \rightarrow {\rm dim}(a) $     \\
%Actor activation  & ${\rm layernorm(Tanh)} $    \\
%&    $\rightarrow {\rm ReLU}  \rightarrow {\rm Tanh}   $  \\
%Critic network (MLP) & ${\rm dim}(s)+64+{\rm dim}(a) $    \\   
% &    $\rightarrow 1024 \rightarrow 1024 \rightarrow 1 $       \\
%Critic activation  & ${\rm layernorm(Tanh)} $  \\
% &   $\rightarrow {\rm ReLU}  $  \\
Agent update frequency & $2$ \\
Target critic network EMA & $0.01$\\
Exploration stddev clip & $0.3$ \\
Exploration stddev value & $0.2$ \\
\hline
Skill combination training steps & $2e+6$ \\
Eval frequency & 10000  \\
Number of Eval episodes & 10 \\
Meta-controller Eval stddev  & $0.2$ \\
%Skill agent eval stddev (Cheetah)   & $0.2$ \\
Skill agent eval stddev  & $0$ \\
\hline
Skill finetuning steps & $1e+5$ \\
Skill choice (random) steps & $4000$ \\
ComSD fixed target skill &  $(0,0.5,0.5,...,0.5)$ \\
Learning rate (Walker\&Quadruped) & $1e-4$\\
Learning rate (Hopper\&Cheetah) & $2e-5$\\
Number of Eval episodes & 10 \\
Eval stddev value   & $0$ \\

\hline
\end{tabular}
}
\label{hyper}
\end{table}

\bibliographystyle{IEEEtran}
\bibliography{tcyb}


% \begin{IEEEbiographynophoto}{Xin Liu, IEEE Graduate student member} %Xin Liu obtained B.Sc. in Computer Science from Shandong University, Jinan, China, and is currently pursuing a Ph.D. degree in Pattern Recognition at the MAIS, Institute of Automation, Chinese Academy of Sciences, Beijing, China.
% %Use $\backslash${\tt{begin\{IEEEbiographynophoto\}}} and the author name as the argument followed by the biography text.

% \begin{IEEEbiographynophoto}{Yaran Chen, IEEE Member}
% %Use $\backslash${\tt{begin\{IEEEbiographynophoto\}}} and the author name as the argument followed by the biography text.

% \begin{IEEEbiographynophoto}{Guixing Chen}
% %Use $\backslash${\tt{begin\{IEEEbiographynophoto\}}} and the author name as the argument followed by the biography text.

% \begin{IEEEbiographynophoto}{Haoran Li, IEEE Member}
% %Use $\backslash${\tt{begin\{IEEEbiographynophoto\}}} and the author name as the argument followed by the biography text.

% \begin{IEEEbiographynophoto}{Dongbin Zhao, IEEE Fellow}
% %Use $\backslash${\tt{begin\{IEEEbiographynophoto\}}} and the author name as the argument followed by the biography text.

\end{document}